\documentclass[preprint,12pt]{elsarticle}

\usepackage{amssymb,amsmath,subfigure}
\usepackage{algorithm}
\usepackage{algpseudocode}
\usepackage{color}

\journal{Information Sciences}

\begin{document}

\begin{frontmatter}



\title{Heterogeneous Federated Learning Systems for Time-Series Power Consumption Prediction with Multi-Head Embedding Mechanism}


\author{Jia-Hao Syu$^a$, Jerry Chun-Wei Lin$^{b,} $\footnote[1]{Corresponding author}, Gautam Srivastava$^{c} $, and Unil Yun$^{d} $\\
$^a$Department of Computer Science and Information Engineering,\\National Taiwan University, Taiwan, f08922011@ntu.edu.tw\\
$^b$Department of Distributed Systems and Informatic Devices,\\Silesian University of Technology, Poland, jerrylin@ieee.org\\
$^c$Department of Math and Computer Science, Brandon University, \\Brandon, Canada, SRIVASTAVAG@brandonu.ca\\
$^d$Department of Computer Science and Engineering, \\Sejong University, South Korea, yunei@sejong.ac.kr}

\begin{abstract}
Time-series prediction is increasingly popular in a variety of applications, such as smart factories and smart transportation.
Researchers have used various techniques to predict power consumption, but existing models lack discussion of collaborative learning and privacy issues among multiple clients.
To address these issues, we propose Multi-Head Heterogeneous Federated Learning (MHHFL) systems that consist of multiple head networks, which independently act as carriers for federated learning.
In the federated period, each head network is embedded into 2-dimensional vectors and shared with the centralized source pool.
MHHFL then selects appropriate source networks and blends the head networks as knowledge transfer in federated learning.
The experimental results show that the proposed MHHFL systems significantly outperform the benchmark and state-of-the-art systems and reduce the prediction error by 24.9\% to 94.1\%.
The ablation studies demonstrate the effectiveness of the proposed mechanisms in the  MHHFL (head network embedding and selection mechanisms), which significantly outperforms traditional federated average and random transfer.
\end{abstract}


\end{frontmatter}

\section{Introduction}     \label{sec:Intro}
Federated learning is an emerging research topic that uses knowledge learned collaboratively by multiple clients (participants) and coordinated by a centralized server (aggregator)~\cite{Fed1}.
It allows clients to learn a machine learning model without sharing private data.
The pioneering algorithms are federated stochastic gradient descent (FedSGD~\cite{FedSGD}) and federated average (FedAvg~\cite{FedAvg}).
The server initializes a global model and broadcasts it to the clients, who train their own models using their local data.
The clients share the model information (gradient or weights) with the server, which aggregates and updates the global model.
After several iterations, clients could incorporate global knowledge and obtain a robust model without sharing local data.

Nevertheless, federated learning in real-world applications faces the challenge of heterogeneity, which has led to the new research topic of heterogeneous federated learning.
It focuses on different client capacities and data distributions.
The different capacities may be related to the computational power of hardware or communication bandwidth~\cite{HFL-IoT,HFL-Adp}, causing the bottleneck in the federated learning process.
Therefore, balancing computation loads between clients is a major branch of heterogeneous learning research, especially in smart building~\cite{New2} and edge computing~\cite{New1}.
Non-IID (Independent and Identically Distributed) data distribution also hinders knowledge sharing in federated learning. 
Acquiring knowledge from datasets with different distributions or even different definitions is the research focus of heterogeneous federation in various filed, such as computer vision~\cite{New3} and smart healthcare~\cite{PDP}.
Heterogeneous federated learning facilitates knowledge sharing and scalability improvement in heterogeneous domains, and is an active research area in machine learning that has significant potential for a variety of applications, such as Smart Homes~\cite{SH}, Smart Factories~\cite{SI}, Smart Grid~\cite{M-TSN}, and Smart Transportations~\cite{ST}.

In the emerging smart transportation industry, the popularity of electric vehicles has led to a growing need for prediction of power consumption to optimize battery performance, minimize energy waste, increase reliability, and support route planning~\cite{EV,EV-PP}.
However, vehicle energy consumption prediction faces serious challenges related to limited energy storage, complex driving conditions, and uncertain environments.
How to make full use of heterogeneous data from different sensors and privacy-conscious users is an important research question.
For example, driving behavior is critical to the power consumption of electric vehicles in smart transportation, but is relevant to user privacy and cannot be directly shared.
Likewise, route and payload information is crucial for the power consumption of autonomous vehicles in smart factories, but is commercially confidential and cannot be directly shared among users (companies).
To address these challenges, researchers have used various techniques, such as convolutional neural networks (CNN) and long-term memory networks (LSTM)~\cite{ENG}, and developed various mechanisms for feature engineering \cite{SMC,NC,TASE}.
However, existing models generally focus on feature engineering from a single data source and lack discussion of collaborative learning and privacy issues among multiple clients.

To achieve the above objectives, we proposed Multi-Head Heterogeneous Federated Learning (MHHFL) systems for time-series power consumption prediction.
The MHHFL has a data preprocessing module for feature packing and pre-classification that facilitates head network embedding and heterogeneous transfer.
The network design of MHHTL consists of multiple head networks and a prediction network, where each head network independently classifies a feature set and acts as a carrier for federated learning.
The classification results are aggregated for the prediction network to make the final time-series power consumption prediction.

In the federated period, the proposed head network embedding mechanism embeds each head network into 2-dimensional (2D) vectors, and clients share the information (weights and embedded vectors) of the head networks with a centralized source pool.
For each (target) head network, the designed selection mechanism selects the appropriate source networks (in the source pool) based on the embedded vectors.
The target head network is then blended with the selected source networks, as knowledge transfer in federated learning.
The advantages and contributions of the proposed MHHFL can be summarized as follows:

\begin{enumerate}
\item The proposed multi-head network and head network embedding mechanism facilitate heterogeneous transmission.
\item The developed selection mechanism and source pool enable asynchronous federated learning in uncertain environments (communication latency).
\item The proposed MHHFL shares only the information of the head networks, which not only increases the efficiency but also reduces privacy leakage~\cite{FedLeakage}.
\item The selection mechanism could filter out malicious clients~\cite{FedMC}, and prevent data poisoning~\cite{FedDP}, and increase the security level of federated learning.
\end{enumerate}

The experimental results show that the proposed MHHFL systems significantly outperform the benchmark and state-of-the-art (SOTA) systems, reducing the prediction error of SOTA by 24.9\% to 94.1\%.
The robustness evaluation is performed to simulate the communication latency, and the MHHFL systems maintain their excellent performance and suppress the SOTA by 4.9\% to 94.5\%.
The ablation studies show the effectiveness of the proposed mechanisms, especially heterogeneous federated learning (head network embedding and selection mechanisms), which significantly outperforms traditional FedAvg and random transfer.

The following sections of the paper are organized as follows.
Section~\ref{sec:LR} surveys the literature on federated learning and power consumption prediction.
Section~\ref{sec:M} presents the proposed MHHFL; the experimental results and discussions are exhibited in Section~\ref{sec:Exp}.
The paper concludes in Section~\ref{sec:Con} with future research directions.
For easy understanding, we summarize the important acronyms and variables in Table~\ref{tab:TAV}.

\section{Literature Review}     \label{sec:LR}
In this section, we first present background knowledge of federated learning in Section~\ref{sec:LR-FL} and review related work on power consumption prediction in Section~\ref{sec:LR-PCP}.

\subsection{Federated Learning}     \label{sec:LR-FL}
Federated learning uses knowledge learned collaboratively by multiple clients (participants) and coordinated by a central server (aggregator)~\cite{Fed1}.
Moreover, federated learning allows multiple clients to learn a machine learning model without sharing private data.
The pioneering federated learning algorithm belongs to federated stochastic gradient descent (FedSGD~\cite{FedSGD}).
The central server first initializes a global model and sends it to (all for selected) clients.
Clients train their own models by stochastic gradient descent on their own local data. Clients randomly select a subset (mini-batch) of local data for training and share the gradients with the server.
The server aggregates and updates the global model based on the obtained gradients and weights (proportional to the batch size) and resends the global model to the clients.
After several iterations, the clients were able to absorb the global knowledge and obtain a robust model without sharing local data.
Shortly after, the now famous and typical algorithm for federated learning, federated average (FedAvg~\cite{FedAvg}), was developed.
FedAvg does not constrain the training algorithms on the clients, but aggregates and averages the model weights to form the global model, which is more general and easier to implement.
In summary, federated learning is a robust algorithm for collaborative learning and knowledge sharing while respecting privacy. This is particularly useful in domains where privacy is a concern (e.g., healthcare~\cite{PDP} and finance~\cite{Fed-Fin}) and in distributed data systems (e.g., Internet of Things, IoT~\cite{Fed-IoT}).

In real-world applications, however, federated learning faces the challenges of heterogeneity, leading to the emerging research topic of heterogeneous federated learning~\cite{HFL}.
Heterogeneous federated learning focuses on two aspects: different capacities and different data distribution among clients.
The different capacities can be either the computing power of the hardware or the bandwidth of the communication.
Cui et al.~\cite{HFL-IoT} proposed a heterogeneous IoT edge federated learning system for the resources-limited scenarios, which requests the clients' resource information, generates client scheduling by linear programming, and allocates resources to the clients.
By real-time monitoring of the resource information and dynamically scheduling the resource, the system optimizes resource and time efficiency.
Zhang et al.~\cite{HFL-Adp} developed adaptive federated learning algorithms for a heterogeneous computing environment.
The algorithms assign adaptive instead of constant workload (training iteration) to clients, which minimizes the runtime gap between clients and maximizes the time efficiency and convergence.

As for heterogeneous data distribution, non-IID data hinders knowledge sharing in federated learning. Moreover, not only the distribution but also the definition of features could be different in realistic scenarios.
Therefore, facilitating knowledge sharing between heterogeneous domains and improving the scalability of federated learning is an active research area in machine learning and holds significant potential for a wide range of applications.
Luo et al.~\cite{New5} proposed an adaptive client sampling algorithm for statistical heterogeneity, which determined the learning time and sampling probabilities to speed up convergence.
Huang et al.~\cite{New3} utilized public data for domain representation and communication through a cross-correlation matrix to facilitate heterogeneous federated learning; Fang and Ye.~\cite{New4} also adopted public data to align heterogeneous models, and further designed a re-weighting scheme to deal with the noise.
Tang et al.~\cite{New6} proposed virtual homogeneous learning to rectify heterogeneous data, which generated virtual homogeneous datasets for federated learning. The virtual datasets were generated by pure noise shared by clients, and had no private information.
Existing methods mainly focus on the external (public or virtual) dataset to select and calibrate the heterogeneous data by statistical approaches.
In this paper, we aim to preserve the intrinsic characteristics of data, and facilitate heterogeneous federated learning through network design and embedding without relying on external datasets.

\subsection{Power Consumption Prediction}     \label{sec:LR-PCP}
Power consumption prediction is an important application for predicting the energy consumption of buildings, facilities, or vehicles to enable smart homes~\cite{SH}, smart factories~\cite{SI}, and smart transportation~\cite{ST}.
Accurate predictions help reduce waste, improve efficiency, optimize management strategies, and detect anomalies.
In the age of artificial intelligence (AI), machine learning algorithms such as regression, data mining, and neural networks are widely used to identify the patterns and relationships between electricity consumption and observed variables.
Therefore, AI-based prediction of electricity consumption is a current research topic for sustainable energy management~\cite{M-SUS,M-CEE} in practical applications.

With sustainable and environmentally friendly characteristics, the popularity of electric vehicles is gradually increasing~\cite{EV}, leading to a growing need for power consumption prediction to optimize battery performance, minimize energy waste, increase reliability, and support route planning~\cite{EV-PP}.
However, power consumption prediction on vehicles faces significant challenges related to limited energy storage, complex driving conditions, and uncertain environments.
How to make full use of data collected from vehicle sensors, weather, and traffic conditions is an important research question related to the use of heterogeneous data.
Kim and Cho~\cite{ENG} adopt traditional CNN and LSTM to predict energy consumption, where the CNN layer is designed to extract information from multiple variables, and the LSTM layer is designed to model temporal information (irregular trends).
Benecki et al.~\cite{SMC} predicted the energy consumption of automated guided vehicles in smart factories.
They developed a systematic feature selection mechanism to deal with heterogeneous data based on the Pearson correlation coefficient (between the prediction features and labels), and used a naive recurrent neural networks for prediction without special network design.
Li et al.~\cite{NC} proposed a multi-view deep neural networks feeding the predefined feature groups into neural networks to extract features, analyze the key factors, and avoid feature interference; Syu et al.~\cite{TASE} further proposed multi-head neural networks with two-stage learning mechanisms to predict time-series information, which independently analyzes and utilizes the feature groups to prevent feature interference.
Existing models for power consumption prediction focus on feature engineering for a single data source, leaving the discussion of collaborative learning and privacy issues between multiple clients.

\section{Proposed Multi-Head Heterogeneous Federated Learning (MHHFL)}     \label{sec:M}
In this paper, we define the research question as time-series power consumption prediction.
Given $NS$ time-series datasets for power consumption, each dataset $j$ has $NF_{j}$ time-series features.
Due to the heterogeneity, the definition and number of features differ among datasets.
For each dataset $j$, the time-series features are denoted as $x_{i,t}$, where $i = \{1, \ldots, NF_{j}\}$ is the index of the feature, and $t$ is the index of time.
Similarly, the power consumption label is denoted as $y_{t}$.
The proposed MHHFL aims to make full use of knowledge in all $NS$ heterogeneous datasets by federated learning and multi-head embedding mechanism.


\subsection{Data Preprocessing}     \label{sec:M-DE}
For each prediction label $y_{t}$, the corresponding time-series feature tensor $X_{t}$ is defined in Eq. (\ref{eq1}) as:
\begin{equation}
\label{eq1}
\\
\begin{bmatrix}
x_{1, t-W} & x_{1, t-W+1} & \ldots & x_{1, t-1} \\
x_{2, t-W} & x_{2, t-W+1} & \ldots & x_{2, t-1} \\
\vdots & \vdots & \ddots & \vdots \\
x_{NF,t-W} & x_{NF,t-W+1} & \ldots & x_{NF,t-1} \\
\end{bmatrix},
\\
\end{equation}
where $W$ is the window size of features ($W \geq 1$), and $X_{t} \in \Bbb R^{NF \times W}$.

To facilitate heterogeneous federated learning in Section~\ref{sec:M-HFL}, a pre-classification mechanism is designed.
The categories for feature tensor $X_{t}$ are denoted as $C_{t} = [\, C_{1,t}, C_{2,t}, \ldots, C_{NF,t} \,]$, where $C_{i,t}$ is defined as:
\begin{equation}   \label{eq:pcm}
\\
C_{i,t} = 
\begin{cases}
~~1 & \text{if}~ x_{i,t} > x_{i,t-1},\\
 -1 & \text{else}.
\end{cases}
\\
\end{equation}

Note that the pre-classification mechanism can be freely designed by the user and could contain information about anomalous events that might be more representative and useful in real implementations.
For example, suddenly increasing values or outliers could be marked as 1; otherwise, they would be marked as -1.
In the following experiments, we define the pre-classification mechanism simply as Eq.~(\ref{eq:pcm}) to represent the increase or non-increase of the feature.

\begin{figure}[th]
    \centering
    \includegraphics[width=1.0\textwidth]{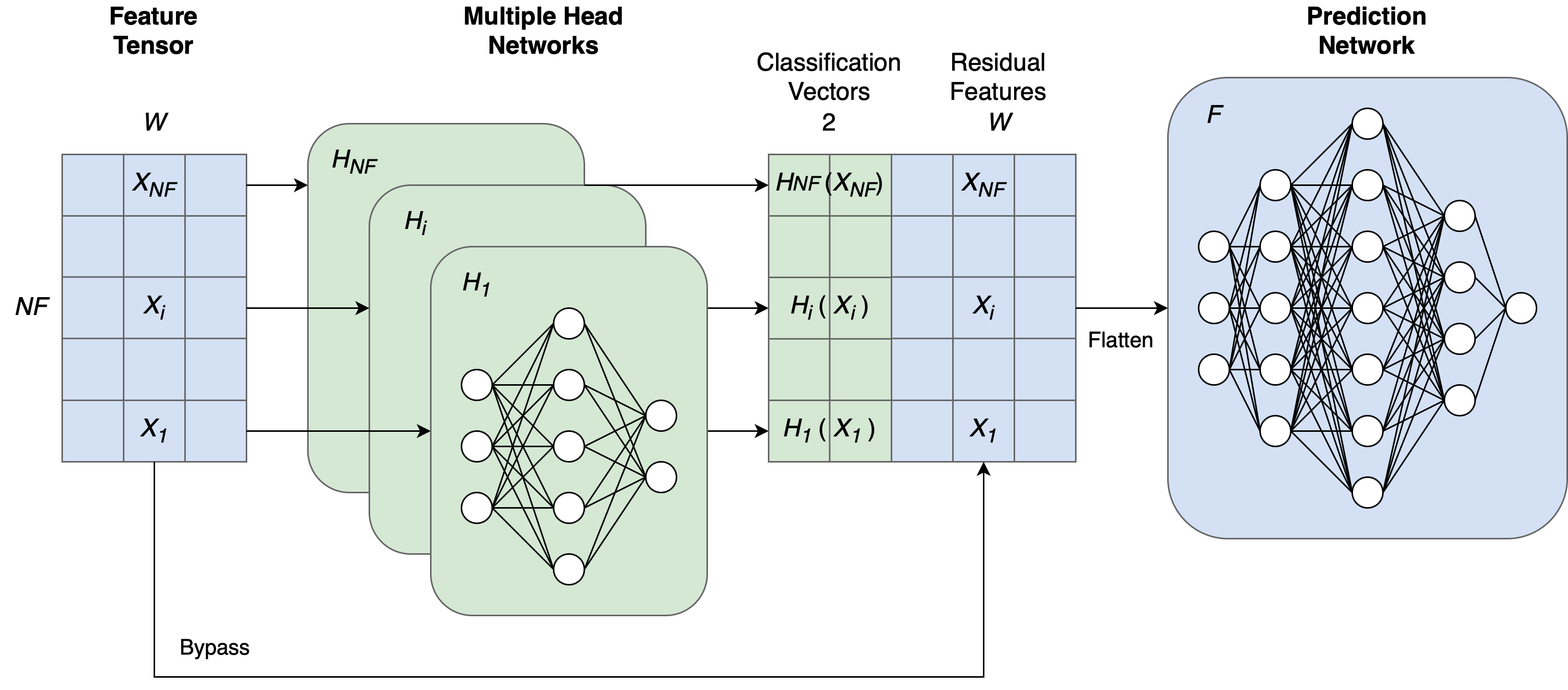}
    \caption{Network design of MHHFL}
    \label{fig:ND}
\end{figure}

\subsection{Network Design}     \label{sec:M-ND}
The network design of MHHFL is referred to~\cite{TITS}, which is exhibited in Figure~\ref{fig:ND}.
The major difference is the architectures of head networks, which are multilayer perceptron (MLP), CNN, LSTM, and Transformer networks~\cite{TITS}, while MHHFL adopts only MLP to demonstrate the generalization and focuses on heterogeneous federated learning.
The detailed architectures of head networks are presented in Section~\ref{sec:Exp-DU}.

The feature tensor $X_{t}$ is first fed into multiple head networks, designed to mitigate data interference~\cite{TASE}, embed the respective features, and facilitate heterogeneous federated learning.
The multiple head networks consist of $NF$ head networks with identical network architecture but different weights, denoted as $H_{1}, H_{2}, \ldots, H_{NF}$.
Each $H_{i}$ adopts the $i^{th}$ row of $X_{t}$ as input, which is the feature vector $X_{i,t} = [\,x_{i,t-W}, x_{i,t-W+1}, \ldots, x_{i,t-1} \,]$; therefore, the input dimension of all $H_{i}$ is $W$.
In addition, the output layer of all $H_{i}$ is a 2-neuron fully connected layer with the softmax activation function, which aims to classify the feature vector into the corresponding $C_{i,t}$, defined in Section~\ref{sec:M-DE}.
In summary, for each feature $i$, the corresponding head network $H_{i}$ takes the $i^{th}$ row of feature tensor $X_{t}$ (feature vector $X_{i,t}$) to classify the binary category of $C_{i,t}$.

Let the 2D classification results $H_{i}(X_{i,t})$ be $p_{i,t} \in \Bbb R^{2}$, and let $p_{t}$ be $[\, p_{1,t}$, $p_{2,t}$, $\ldots$, $p_{NF,t} \,]$.
The classification vectors, $p_{t}$, are then contacted with the origin input tensor $X_{t}$ as the input for the prediction network $F$.
The final prediction of $y_{t}$ is expressed as:
\begin{equation}
\\
y'_{t} = F( X_{t}, p_{t} ).
\\
\end{equation}

\subsection{Heterogeneous Federated Learning}     \label{sec:M-HFL}
Suppose there are $NS$ clients (source domains) in heterogeneous federated learning, and each client has its own features and neural networks (defined in Section~\ref{sec:M-ND}).
The number, meaning, and distribution of features in each source domain may be different, leading to heterogeneity in federated learning.
Let the number of features in the $j^{th}$ source domain be $NF_{j}$, where $j = \{1, \ldots, NS \}$.
Moreover, the $i^{th}$ head network in the $j^{th}$ source domain is denoted as $H^{j}_{i}$.

\begin{figure}
    \centering
    \includegraphics[width=0.8\textwidth]{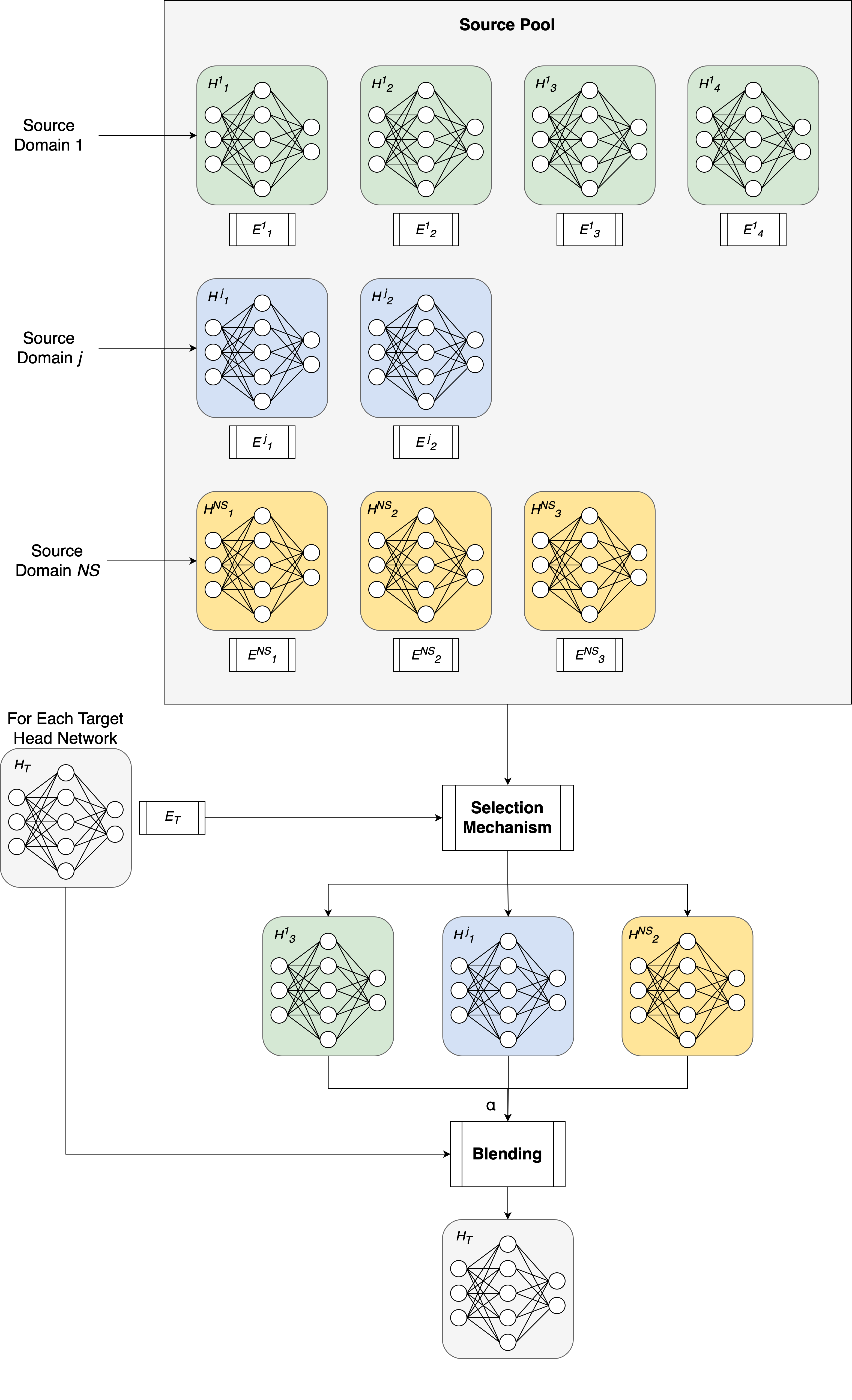}
    \caption{Proposed heterogeneous federated learning}
    \label{fig:HFL}
\end{figure}

We illustrate the flowchart of the proposed heterogeneous federated learning in Figure~\ref{fig:HFL}.
Note that there are $NS$ domains participating in the federated learning, and we only draw 3 domains for illustrative purposes in Figure~\ref{fig:HFL}.
For each federated period (detailed in Section~\ref{sec:M-HFL-OV}), each source domain computes the embedded vectors of the head networks and shares the weights and embedded vectors of the head networks with the centralized source pool (detailed in Section~\ref{sec:M-HFL-HNE}).
For each target head network (of each domain), the proposed selection mechanisms then select the appropriate source head networks based on the embedded vectors and blend the networks as update processes in federated learning (detailed in Section~\ref{sec:M-HFL-SM}).

\subsubsection{Head Network Embedding}     \label{sec:M-HFL-HNE}
We adopt the concept of pulling-pushing forces~\cite{PullPush} to embed head networks and assist heterogeneous feature selection.
Following Section~\ref{sec:M-ND}, the head network $H^{j}_{i}$ takes the feature vector $X_{i,t}$ to classify the binary category of $C_{i,t}$.
Let the weights of the output layer of $H^{j}_{i}$ be $w \in \Bbb R^{d \times 2}$, and $w_1$ and $w_2$ represent the first and second columns of $w$.
$d$ is the input dimension of the output layer, and the input (of the output layer) is denoted as $h_{t} \in \Bbb R^{d}$.
The softmax activation function then calculates the probabilities of classification as:
\begin{equation}
\begin{aligned}
\\
p_{i,t,1} &= \frac{ e^{h_{t} w_1} }{ e^{h_{t} w_1} + e^{h_{t} w_2} },\\
\\
p_{i,t,2} &= \frac{ e^{h_{t} w_2} }{ e^{h_{t} w_1} + e^{h_{t} w_2} }.
\\
\end{aligned}
\end{equation}
Cross-entropy is adopted~\cite{PullPush} to calculate the loss, $L$, as:
\begin{equation}
\begin{aligned}
\\
L &= \sum_{t, C_{i,t}=1}  \log p_{i,t,1} ~+~ \sum_{t, C_{i,t}=-1} \log p_{i,t,2},\\
  &= \sum_{t, C_{i,t}=1}  \log \frac{ e^{h_{t} w_1} }{ e^{h_{t} w_1} + e^{h_{t} w_2} } 
 ~+~ \sum_{t, C_{i,t}=-1} \log \frac{ e^{h_{t} w_2} }{ e^{h_{t} w_1} + e^{h_{t} w_2} }.
\\
\end{aligned}
\end{equation}
Let $D$ be $e^{h_{t} w_1} + e^{h_{t} w_2}$, and the gradient of $w_1$ is derived as:
\begin{equation}     \label{eq:Grad}
\begin{aligned}
\\
\frac{\partial L}{\partial w_1} = 
&\sum_{t, C_{i,t}=1} \frac{ D }{ e^{h_{t} w_1} }~ ( \frac{ e^{h_{t} w_1} }{ D }  )' ~+~ \sum_{t, C_{i,t}=-1} \frac{ D }{ e^{h_{t} w_2} }~ ( \frac{ e^{h_{t} w_2} }{ D }  )'
\\
\\
= &\sum_{t, C_{i,t}=1}  \frac{ D }{ e^{h_{t} w_1} }~ \frac{ e^{h_{t} w_1} h_{t} D ~-~ e^{h_{t} w_1} e^{h_{t} w_1} h_{t} }{ D^{2} } ~+~\\
  &\sum_{t, C_{i,t}=-1} \frac{ D }{ e^{h_{t} w_2} }~ \frac{ - e^{h_{t} w_2} e^{h_{t} w_1} h_{t} }{ D^{2} }
\\
\\
= &\sum_{t, C_{i,t}=1}  \frac{  h_{t} (D ~-~ e^{h_{t} w_1}) }{ D } + 
   \sum_{t, C_{i,t}=-1} \frac{ -h_{t} e^{h_{t} w_1} }{ D }
\\
\\
= &\sum_{t, C_{i,t}=1}  \Bbb~ h_{t} (1 - p_{i,t,1}) + 
   \sum_{t, C_{i,t}=-1} \Bbb~ h_{t} (  - p_{i,t,1}).
\\
\end{aligned}
\end{equation}

Referred from~\cite{PullPush}, we define the pulling force of $w_1$ as:
\begin{equation}
\\
\frac{ \sum_{t, C_{i,t}=1}  \text{Avg}(h_{t}) (1 - p_{i,t,1}) }{ T },
\\
\end{equation}
where Avg$(\cdot)$ is the average function of a vector, and $T$ is the number of sample $t$ (in the batch).
Therefore, the pulling force is the average value of the first terms of Equation~(\ref{eq:Grad}).
Likewise, the pushing of $w_1$ is defined as:
\begin{equation}
\\
\frac{ \sum_{t, C_{i,t}=-1} \text{Avg}(h_{t}) (  - p_{i,t,1}) }{ T },
\\
\end{equation}
which is the average value of the second term of Equation~(\ref{eq:Grad}).
Similarly, the pulling and pushing forces of $w_2$ are:
\begin{equation}
\\
\frac{ \sum_{t, C_{i,t}=1}  \text{Avg}(h_{t}) (  - p_{i,t,2}) }{ T },~
\frac{ \sum_{t, C_{i,t}=-1} \text{Avg}(h_{t}) (1 - p_{i,t,2}) }{ T }.
\\
\end{equation}
The two pairs ($w1$ and $w2$) of pulling-pushing forces are the determination basis for heterogeneous federated learning.
For easy representation,  $E^{j}_{i} \in \Bbb R^{4}$ represents the two-pair force of the head network $H^{j}_{i}$ with the dataset in domain $j$.

The above head network embedding mechanism is based on gradient, while Syu et al.~\cite{TITS} proposed a data-based head network embedding mechanism, and the embedded forces are defined as:
\begin{equation}
\\
E^{j}_{i} =  \left[
\begin{aligned}
&\frac{ \sum_{t, C_{i,t}=1} ~ \text{Avg}(w_{1}) (1 - p_{i,t,1}) }{ T },~
 \frac{ \sum_{t, C_{i,t}=-1}~ \text{Avg}(w_{1}) (  - p_{i,t,1}) }{ T }~
 \\
&\frac{ \sum_{t, C_{i,t}=1} ~ \text{Avg}(w_{2}) (  - p_{i,t,2}) }{ T },~
 \frac{ \sum_{t, C_{i,t}=-1}~ \text{Avg}(w_{2}) (1 - p_{i,t,2}) }{ T }~
 \end{aligned}
\right].
\\
\end{equation}


For each federated learning period, each source domain $j$ will share both the embedded vectors and weights of all head networks to the source pool, which are $E^{j}_{i}$ and $H^{j}_{i}$ for $j = \{1, \ldots, NS \}$ and $i = \{1, \ldots, NF_{NS}\}$, respectively.
However, due to various communication errors and delays, each source domain $j$ has a probability $DP$ that does not share any knowledge during the federated learning process, and the domain-embedding vectors and weights will remain the same as the previous versions.

\subsubsection{Selection and Blending Mechanisms}     \label{sec:M-HFL-SM}
In each federated learning period, after sharing local knowledge (Section~\ref{sec:M-HFL-HNE}), each source domain will use the global knowledge in the source pool to update the local networks.
For each head network $i$ in each source domain $j$, the most appropriate head networks (in the source pool) are selected based on the L2 distance of the embedded vectors, and the network weights are mixed to absorb the global and heterogeneous knowledge.
We developed two selection mechanisms for MHHFL, namely multiple selection and single selection.

The multiple-selection mechanism is illustrated in Figure~\ref{fig:HFL}.
Given a target head network $H^{j}_{i}$, the multiple-selection mechanism selects the most suitable head network $H^{*l}$ from each source domain ($l = \{1, \ldots, NS \}$), and is denoted as:
\begin{equation}
\\
H^{*l} = \operatorname*{arg\,min}_{H^{l}_{k}, ~k=1, \ldots, NF_{l}} \text{Avg}( (E^{l}_{k} - E^{j}_{i})^{2} ).
\\
\end{equation}
The mechanism then updates (blends) the target networks weights by the selected head networks, and defines the blending scale $B$ by the softmax function as:
\begin{equation}
\\
B = \text{softmax}( ~\left[ \text{Avg}( (E^{*l} - E^{j}_{i})^{2} ) ~|~ l = 1, \ldots, NS \right]~ ),
\\
\end{equation}
where $E^{*l}$ is the embedded vector of the corresponding selected head network $H^{*l}$, and $B \in \Bbb R^{NS}$ with the summation of 1.
The mechanism then updates target networks $H^{j}_{i}$ to:
\begin{equation}
\\
H^{j}_{i} = (1-\alpha) H^{j}_{i} + \alpha \sum_{l=1}^{NS} B_{l} H^{*l},
\\
\end{equation}
where $\alpha$ is the scale of the blending.

The single-selection mechanism is a simplified version that only selects one suitable head network in the source pool.
The most suitable head network $H^{*}$ is defined as:
\begin{equation}
\\
H^{*} = \operatorname*{arg\,min}_{H^{l}_{k}, ~l=1, \ldots, NS, ~k=1, \ldots, NF_{l}} \text{Avg}( (E^{l}_{k} - E^{j}_{i})^{2} ),
\\
\end{equation}
and the target network $H^{j}_{i}$ is updated to:
\begin{equation}
\\
H^{j}_{i} = (1-\alpha) H^{j}_{i} + \alpha H^{*}.
\\
\end{equation}
We will compare the effectiveness of multiple- and single-selection mechanisms in the following experiments.

\begin{algorithm}[t]
\footnotesize 
  \caption{Proposed MHHFL}
  \label{alg:MHHFL}
  \hspace*{\algorithmicindent} \textbf{Input:} Dataset of the domain\\
  \hspace*{\algorithmicindent} \textbf{Output:} Trained network and testing results\\
  \begin{algorithmic}[1]
    \State Divide the dataset into training, validation, and testing sets
    \State Pack the feature tensor and pre-classification the features     \Comment{Section~\ref{sec:M-DE}}
    \State Initialize prediction networks for each source domain ($F_{j}$)  \Comment{Section~\ref{sec:M-ND}}
    \State Initialize head networks for each source domain and each feature ($H^{j}_{i}$)     \Comment{Section~\ref{sec:M-ND}}
    \State Pretrain networks by training and validation sets
    \For{ each training epoch}     \Comment{training}
        \For{each training batch}
            \If{the federated period}      \Comment{heterogeneous federated learning}
                \State Calculate the embedded vectors of all head networks  \Comment{Section~\ref{sec:M-HFL-HNE}}
                \For{each source domain}
                    \If{random int $> DR$}
                        \State Share the local knowledge with the source pool
                    \EndIf
                \EndFor
                \State Execute the selection mechanism    \Comment{Section~\ref{sec:M-HFL-SM}}
                \State Execute the blending mechanism to update head networks    \Comment{Section~\ref{sec:M-HFL-SM}}
            \EndIf
            \State Make predictions
            \State Train the networks
        \EndFor
        \State Evaluate the validation set (same as lines 7 to 20)
        \State Save the networks if the validation loss is enhanced
    \EndFor
    \State Evaluate the testing set (same as lines 7 to 20)
    \State \Return Trained network and testing results
  \end{algorithmic}
\end{algorithm}

\subsubsection{MHHFL Overview}     \label{sec:M-HFL-OV}
In this subsection, we summarize the proposed MHHFL in Algorithm~\ref{alg:MHHFL}, and summarize the important acronyms and variables in Table~\ref{tab:TAV} for easy understanding.
MHHFL splits the dataset into a training set, a validation set, and a testing set (line 1), and data preprocessing is performed to pack the feature tensors and pre-classification the features (line 2), as mentioned in Section~\ref{sec:M-DE}.
For each domain (client) $j$, initialize the prediction networks $F_{j}$ and the head networks $H^{j}_{i}$, where $j = \{1, \ldots, NS \}$, $i = \{1, \ldots, NF_{j} \}$ (lines 3 and 4), as designed in Section~\ref{sec:M-ND}.
To acquire sufficient knowledge for sharing, we train the networks using training data (traditional training process without federated learning) with the save-best mechanism on the validation set (line 5).

\begin{table}[h]
\caption{Table of Acronyms and Variables}
\centering
\begin{tabular}{|l|l|}
\hline
\textbf{Acronyms} & \textbf{Description} \\
\hline
2D & 2-Dimensional\\
\hline
MHHFL & Multi-Head Heterogeneous Federated Learning \\
\hline
MHHFL-S & MHHFL with the single selection mechanism\\
\hline
MHHFL-M & MHHFL with the multiple selection mechanism\\
\hline
MHHFL-G & MHHFL with the gradient-based embedding mechanism\\
\hline
MHHFL-D & MHHFL with the data-based embedding mechanism\\
\hline
\multicolumn{2}{}{} \\
\hline
\textbf{Variable} & \textbf{Description} \\
\hline
$NS$ & Number of source domains \\
\hline
$NF$ & Number of features \\
\hline
$W$ & Window size of features vector and tensor, $W \geq 1$\\
\hline
$y_{t}$ & Label at time $t$\\
\hline
$x_{i,t}$ & Feature $i$ at time $t$\\
\hline
$X_{i,t}$ & Feature vector, $[\,x_{i,t-W}, x_{i,t-W+1}, \ldots, x_{i,t-1} \,]$\\
\hline
$X_{t}$ & Feature tensor, $[\,X_{1,t}, X_{2,t}, \ldots, X_{NF,t} \,]$\\
\hline
$C_{i,t}$ & Categories for feature vector $X_{i,t}$\\
\hline
$C_{t}$ & Categories for feature tensor $X_{t}$, $[\,C_{1,t}, C_{2,t}, \ldots, C_{NF,t} \,]$\\
\hline
$F$& Prediction network \\
\hline
$H_{i}$ & Head network for feature $i$\\
\hline
$p_{i,t}$ & Classification results of $H_{i}(X_{i,t})$\\
\hline
 $p_{t}$ & $[\, p_{1,t}$, $p_{2,t}$, $\ldots$, $p_{NF,t} \,]$\\
\hline
$E^{j}_{i}$ & Embedded vectors of head network $H^{j}_{i}$, $E^{j}_{i} \in \Bbb R^{4}$\\
\hline
$B$ & Blending scale for federated learning\\
\hline
$\alpha$ & Scale of the blending\\
\hline
$DR$ & Probability that a source domain does not share knowledge\\
\hline
\end{tabular}
\label{tab:TAV}
\end{table}

In the training process, the networks first make the prediction for each batch (line 18) and update the networks based on the losses (line 19).
Note that there are $1+NF$ losses for a domain, where one loss is the final prediction error (mean square error, MSE, for prediction network $F$) and $NF$ losses are classification errors (cross-entropy loss for head networks $H^{j}_{i}$, $i = \{1, \ldots, NF \}$).
If the batch is a federated period (set to every ten batches in the following experiments), the proposed heterogeneous federated learning is performed before prediction and update (lines 8 to 17).
The embedded vectors of each head network of each domain are first computed (line 9), as described in Section~\ref{sec:M-HFL-HNE}.
For each source domain, there is a probability of $1 - DR$ that the domain shares local knowledge with the source pool, including the embedded vectors and weights of all head networks (lines 10 to 14).
After the sharing process, each head network runs the selection mechanism to choose the most appropriate head networks for mixing and acquiring global knowledge (lines 15 to 16), as shown in Section~\ref{sec:M-HFL-SM}.
Note that the batch sizes for the source domains are the same, but the number of data instances may be different. Therefore, some clients will stop passing data to the source pool, while other clients will continue to train and perform heterogeneous federated learning.

After a training epoch, the validation set is used to validate the current networks (line 21), and the save-best mechanism is applied to save the networks with the lowest prediction error (line 22).
After the training and validation are completed, the testing set is used to show the performance of the proposed MHHFL (line 24).
Note that the validation and testing process is the same as the training process described in lines 7 to 20 of Algorithm~\ref{alg:MHHFL}.

Compared to traditional federated learning algorithms, we summarized the strengths of the proposed MHHFL, including heterogeneity, asynchrony, efficiency, privacy, and security.
The multi-head network facilitates knowledge sharing and acquisition from heterogeneous source domains, and the designed source pool enables asynchronous federated learning that might be hindered by client communication latency and duration.
Moreover, traditional federated learning algorithms achieve a certain level of privacy by sharing models instead of data; however, they still face the risk of privacy leakage~\cite{FedLeakage}, as data distribution can potentially be inferred through reverse engineering from model (gradient) sharing.
The proposed MHHFL system improves upon this by sharing only head network information, which not only increases efficiency but also mitigates privacy leakage by reducing the scale of model sharing.
Last but not least, the selection mechanism can filter out malicious clients~\cite{FedMC} and prevent data poisoning~\cite{FedDP}, which increases the security level of federated learning.

As for the computational complexity, let the average number of features (among $NS$ source domains) be $\overline{NF}$, and let the length of data be $T$.
Suppose the network size is proportional to the number of features. In that case, the computational complexities of conventional federated learning algorithms (FedAvg and FedSGD) are $O(T \cdot NS \cdot \overline{NF})$; otherwise, they are $O(T \cdot NS)$.
As for the proposed MHHFL, the computational complexity of the embedding and selecting mechanisms are $O(T \cdot NS \cdot \overline{NF})$, and the complexity of the blending mechanism is $O(T \cdot NS)$.
Therefore, the overall computational complexity of the proposed MHHFL is $O(T \cdot NS \cdot \overline{NF})$, which is equivalent to the conventional federated learning algorithms (if the network size is proportional to the number of features).

\section{Experimental Results}     \label{sec:Exp}
In this section, we first present the data usage, benchmark system, and experimental setting in Section~\ref{sec:Exp-DU}.
The prediction evaluation and robustness evaluation are then presented in Sections~\ref{sec:Exp-PE} and~\ref{sec:Exp-RO}.
Finally, the ablation studies and sensitivity analysis are performed in Sections~\ref{sec:Exp-AS} and~\ref{sec:Exp-SA}.

\subsection{Data Usage and Benchmark Systems}     \label{sec:Exp-DU}
In this work, we use two datasets with four heterogeneous domains for power consumption prediction.
The first dataset is AIUT, a real-world dataset of automated ground vehicles in the smart factory collected in 2022 by AIUT Co., Ltd., Gliwice, Poland.
The AIUT dataset consists of 10 hertz (HZ) information on motor, control modes, and battery for 12 trials with a total of 120,000 instances.
The second dataset is Husky~\cite{Husky}, a public dataset of the Husky A200 unmanned ground vehicle collected by Carnegie Mellon University in 2021.
The Husky dataset consists of 10 HZ information on motor, location, and battery, and contains datasets for four different routes called HuskyA, HuskyB, HuskyC, and HuskyD.
However, HuskyD does not have enough datasets for training, and we ignore HuskyD in the following experiments.
HuskyA contains 35 experiments with 131,987 instances; HuskyB contains 35 experiments with 73,545 instances; HuskyB contains 17 experiments with 23,342 instances.
Both datasets contain many features, and we choose the same features as~\cite{SMC,TASE} to predict power consumption, with the AIUT and Husky datasets obtaining 25 and 19 features, respectively.
In summary, we obtain four heterogeneous source domains for the following experiments, including AIUT, HuskyA, HuskyB, and HuskyC.
We also split the dataset into 60\% for training, 20\% for validation, and 20\% for testing, based on the number of experiments.

We use four existing power consumption prediction systems for comparison, including SMC~\cite{SMC}, NC~\cite{NC}, ENG~\cite{ENG}, and TASE~\cite{TASE}, presented in Section~\ref{sec:LR-PCP}.
For a fair comparison, we simply adjust the number of neurons so that the systems have a similar number of parameters listed in Table~\ref{tab:para}.
Note that different datasets have different numbers of features, which affects the number of parameters of the neural networks in each system.
All systems are trained with mean square error loss functions, the save-best mechanism, the Adam optimizer, learning rates of 0.01, a stack size of 600, and a window size $W$ of 5.
The four benchmark systems are trained for 100 epochs, while the proposed MHHFL is trained for only 25 epochs because MHHFL performs federated learning on four datasets ($25 \times 4 = 100$).
As for the proposed MHHFL, the federated period is defined as every ten batches (10 minutes $= 10$ batch $\times 600$ instances per batch $\times 10$ HZ), and the details of the network design are shown in Table~\ref{tab:ND}, where LRelu is the Leaky ReLU activation function.
The hyperparameters $DR$ (probability that a source domain does not share knowledge, defined in Section~\ref{sec:M-HFL-OV}) and $\alpha$ (scale of the blending, defined in Section~\ref{sec:M-HFL-SM}) are set to 0.5 and 0.1, respectively.

\begin{table}[th]
\footnotesize 
\caption{Number of Parameters}
\label{tab:para}
\centering
\begin{tabular}{|l|r|r|r|r|r|r|}
\multicolumn{5}{}{} \\
\hline
& SMC~\cite{SMC} & NC~\cite{NC} & ENG~\cite{ENG} & TASE~\cite{TASE} & MHHFL\\
\hline
AIUT    & 70,633 & 71,715 & 74,801 & 75,500 & 78,987 \\
\hline
Hysky   & 68,905 & 62,391 & 70,481 & 68,750 & 60,735 \\
\hline
Average & 69,769 & 67,053 & 72,641 & 72,125 & 69,861 \\
\hline
\end{tabular}
\end{table}

\begin{table}[th]
\footnotesize 
\caption{Details of Network Design}
\label{tab:ND}
\centering
\begin{tabular}{|c|c|c|c|c|c|c|c|}
\multicolumn{8}{}{} \\
\hline
\multicolumn{4}{|c|}{Head Network ($H$)} & \multicolumn{4}{c|}{Prediction Network ($F$)} \\
\hline
Layer & Input & Ouput & Activation & Layer & Input & Ouput & Activation \\
\hline
Linear & $W$ &  8 & LReLU   & Linear & $7 NF$ &  8 & LReLU \\
Linear &   8 & 64 & LReLU   & Linear &   8 & 64 & LReLU \\
Linear &  64 & 32 & LReLU   & Linear &  64 & 32 & LReLU \\
Linear &  32 &  8 & LReLU   & Linear &  32 &  8 &  \\
Linear &   8 &  2 & Softmax & Linear &   8 &  1 &  \\
\hline
\end{tabular}
\end{table}

\begin{table}[th]
\footnotesize 
\caption{Prediction Evaluation}
\label{tab:PE}
\centering
\begin{tabular}{|l|r|r|r|r|c|}
\multicolumn{6}{}{} \\
\hline
\multicolumn{6}{|c|}{\textbf{Validation MSE}} \\
\hline
\multicolumn{1}{|c|}{\textbf{Model}} & \multicolumn{1}{c|}{\textbf{AIUT}} & \multicolumn{1}{c|}{\textbf{HuskyA}} & \multicolumn{1}{c|}{\textbf{HuskyB}} & \multicolumn{1}{c|}{\textbf{HuskyC}} & \multicolumn{1}{c|}{\textbf{Avg. Rank}}\\
\hline
SMC~\cite{SMC}   &  3,159 (7) & 6,218 (7) & 5,484 (8) & 70,180 (8) & (7.5) \\
NC~\cite{NC}     & 21,485 (8) & 2,010 (6) & 4,690 (7) & 56,229 (7) & (7.0) \\
ENG~\cite{ENG}   & \textbf{1,551 (1)} & 6,296 (8) & 2,762 (6) & 39,998 (6) & (5.3) \\
TASE~\cite{TASE} &  2,573 (5) & 1,588 (5) & 1,449 (5) & 21,530 (5) & (5.0) \\
MHHFL-SG         &  2,437 (3) & 1,254 (3) & \textbf{1,153 (1)} & 16,072 (4) & (2.8) \\
MHHFL-SD         &  2,787 (6) & 1,251 (2) & 1,170 (3) & \textbf{15,995 (1)} & (3.0) \\
MHHFL-MG         &  2,535 (4) & 1,261 (4) & 1,180 (4) & 16,039 (3) & (3.8) \\
MHHFL-MD         &  2,253 (2) & \textbf{1,239 (1)} & 1,165 (2) & 16,010 (2) & \textbf{(1.8)} \\
\hline
\multicolumn{6}{}{} \\
\hline
\multicolumn{6}{|c|}{\textbf{Testing MSE}} \\
\hline
\multicolumn{1}{|c|}{\textbf{Model}} & \multicolumn{1}{c|}{\textbf{AIUT}} & \multicolumn{1}{c|}{\textbf{HuskyA}} & \multicolumn{1}{c|}{\textbf{HuskyB}} & \multicolumn{1}{c|}{\textbf{HuskyC}} & \multicolumn{1}{c|}{\textbf{Avg. Rank}}\\
\hline
SMC~\cite{SMC}   &   625,640 (7) & 5,968 (7) & 3,710 (8) & 71,105 (8) & (7.5) \\
NC~\cite{NC}     & 8,809,772 (8) & 3,122 (6) & 3,381 (7) & 64,311 (7) & (7.0) \\
ENG~\cite{ENG}   &   375,824 (5) & 5,978 (8) & 2,384 (6) & 39,794 (6) & (6.3) \\
TASE~\cite{TASE} &   483,268 (6) & 1,966 (5) & 1,105 (5) & 20,972 (5) & (5.3) \\
MHHFL-SG         &    30,976 (3) & 1,489 (4) &   618 (2) & 14,047 (4) & (3.3) \\
MHHFL-SD         &    28,287 (2) & 1,477 (2) & \textbf{606 (1)} & 13,460 (2) & \textbf{(1.8)} \\
MHHFL-MG         & \textbf{17,238 (1)} & 1,488 (3) &   637 (3) & 13,533 (3) & (2.5) \\
MHHFL-MD         &    32,506 (4) & \textbf{1,467 (1)} &   638 (4) & \textbf{13,427 (1)} & (2.5) \\
\hline
\end{tabular}
\end{table}

\subsection{Prediction Evaluation}     \label{sec:Exp-PE}
In this section, we compare the prediction performance of the systems in Table~\ref{tab:PE}, where the values in parentheses represent the rankings among the columns (systems) and Avg. Rank represents the average ranking.
Additionally, we bold the best-performing values on each dataset (among the columns).
For the proposed MHHFL, the symbols S and M represent the single and multiple selection mechanisms defined in Section~\ref{sec:M-HFL-SM}, and the symbols G and D represent the gradient-based and data-based embedding mechanisms defined in Section~\ref{sec:M-HFL-HNE}.

In Table~\ref{tab:PE}, it can be seen that the proposed MHHFL systems (MHHFL-SG, -SD, -MG, and -MD) always achieve the lowest validation error on the Husky datasets, and the system MHHFL-MD achieves the best average ranking of 1.8.
For the testing sets, the results are similar to the validation, and the four proposed MHHFL systems always achieve the lowest prediction error on all datasets, and the system MHHFL-SD achieves the best average ranking of 1.8.
The experimental results exhibit the effectiveness and significant improvements of the proposed MHHFL systems; among them, the data-based embedding mechanism (D) obtains better performance than the gradient-based embedding mechanism (G).

As for the benchmark systems, SMC~\cite{SMC} adopts only naive recurrent neural networks, and the worst performance with an average ranking of 7.5.
NC~\cite{NC} and ENG~\cite{ENG} utilize the multi-view neural networks, CNN, and LSTM, and provide advanced designs in the neural networks.
NC~\cite{NC} and ENG~\cite{ENG} obtain the average ranking of 7.0 and 6.3 on the testing sets, which perform better than SMC~\cite{SMC}.
TASE~\cite{TASE} design two-stage learning mechanisms with feature groups, and further improve the performance with an average ranking of 5.3 on the testing sets.
Among the benchmark systems, the SOTA (best-performing) system belongs to TASE~\cite{TASE}, which achieves an average of 5.0 and 5.3 on the validation and test datasets, respectively; however, is significantly underperformed by the proposed MHHFL systems.
Comparing the proposed MHHFL-SD with the SOTA system, MHHFL-SD reduces the MSE of the tests by 94.1\% (1 - 28,287 / 483,268) for AIUT, 24.9\% (1 - 1,477 / 1,966) for HuskyA, 45.2\% (1 - 606 / 1,105) for HuskyB, and 35.8\% (1 - 13,460 / 20,972) for HuskyC.
In summary, all the proposed MHHFL systems significantly outperform the benchmark systems and reduce the prediction error of SOTA by 24.9\% to 94.1\%.

\begin{figure}
    \centering
    \subfigure[AIUT]{  \includegraphics[width=0.4\textwidth]{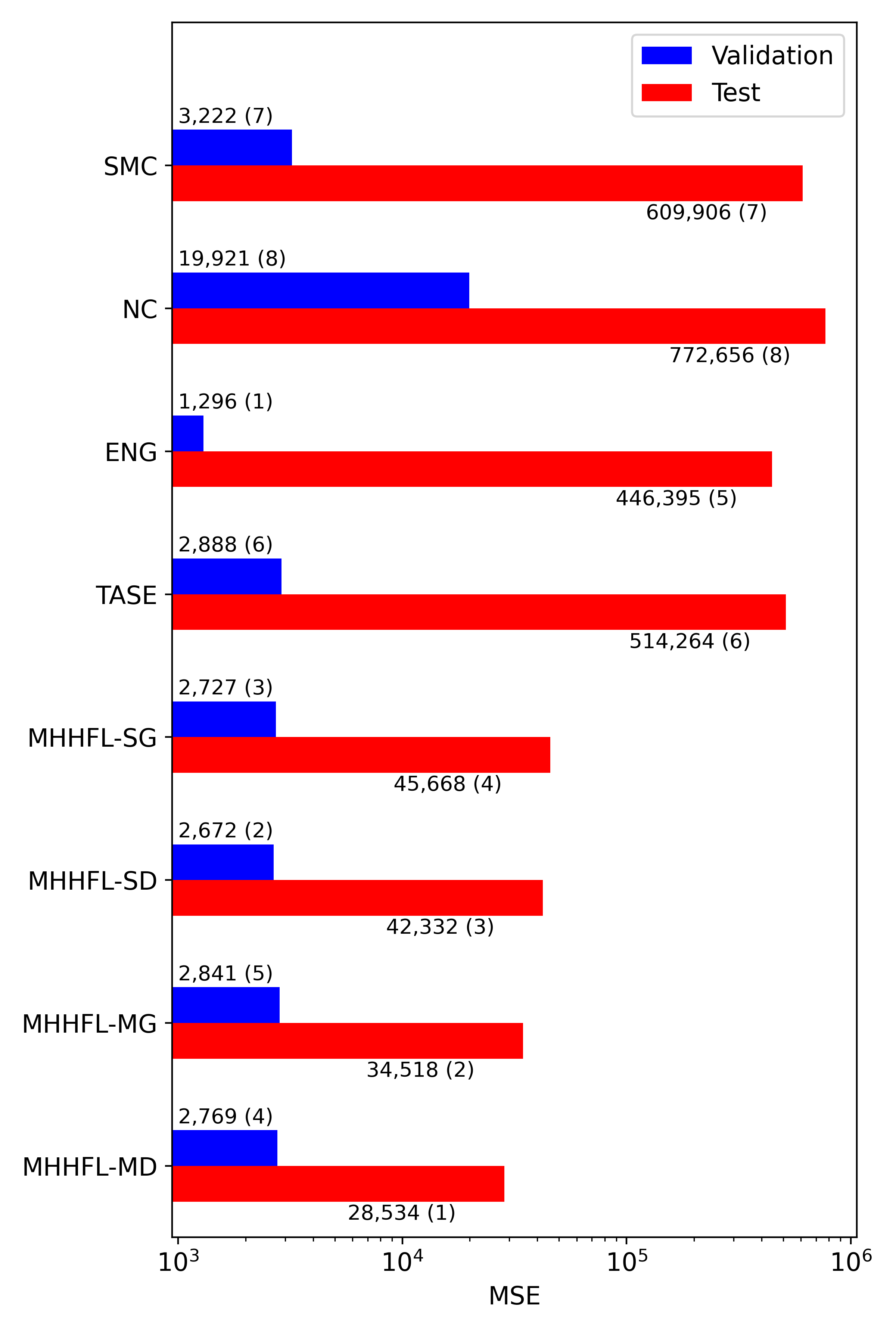}} 
    \subfigure[HuskyA]{\includegraphics[width=0.4\textwidth]{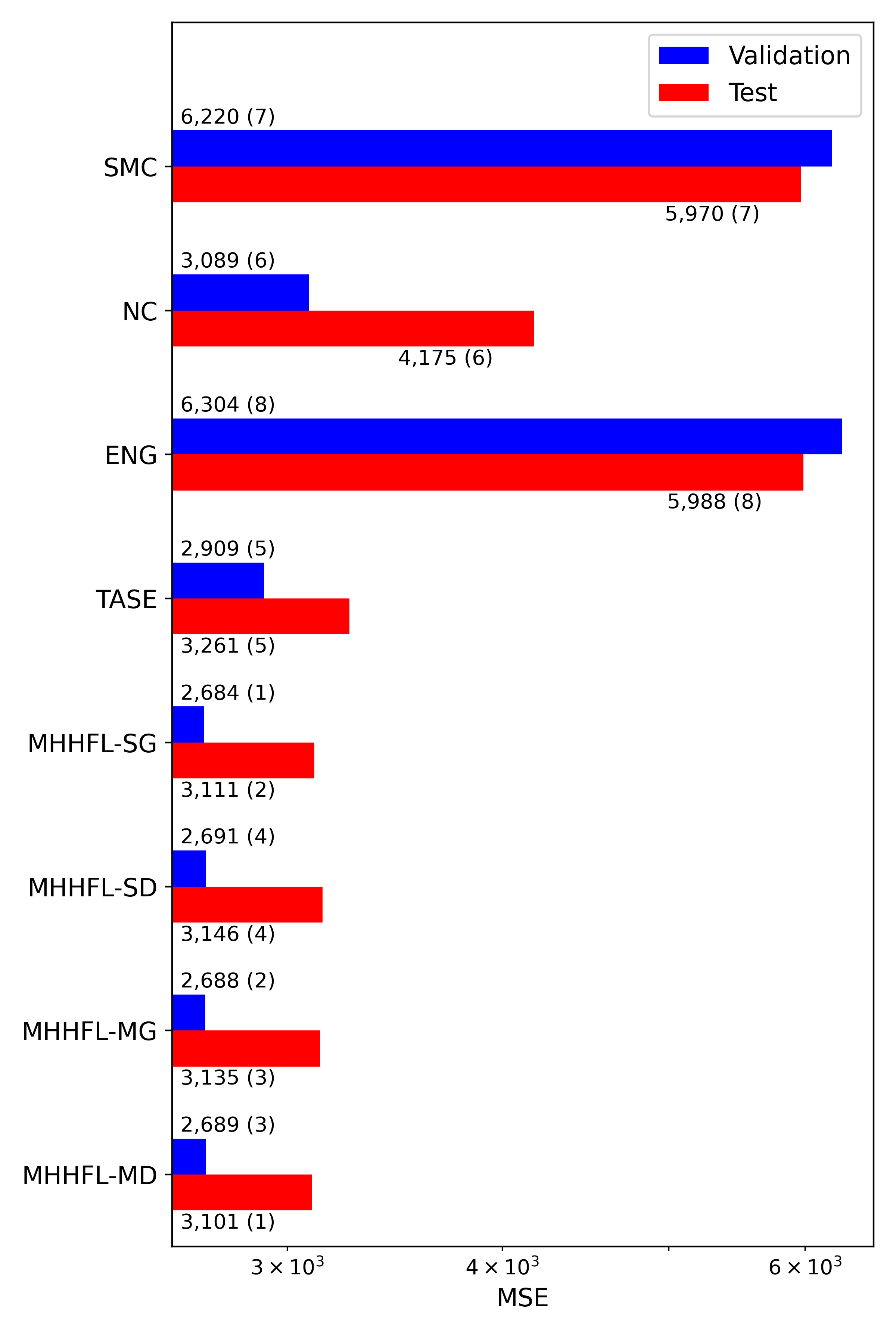}} 
    \centering
    \subfigure[HuskyB]{\includegraphics[width=0.4\textwidth]{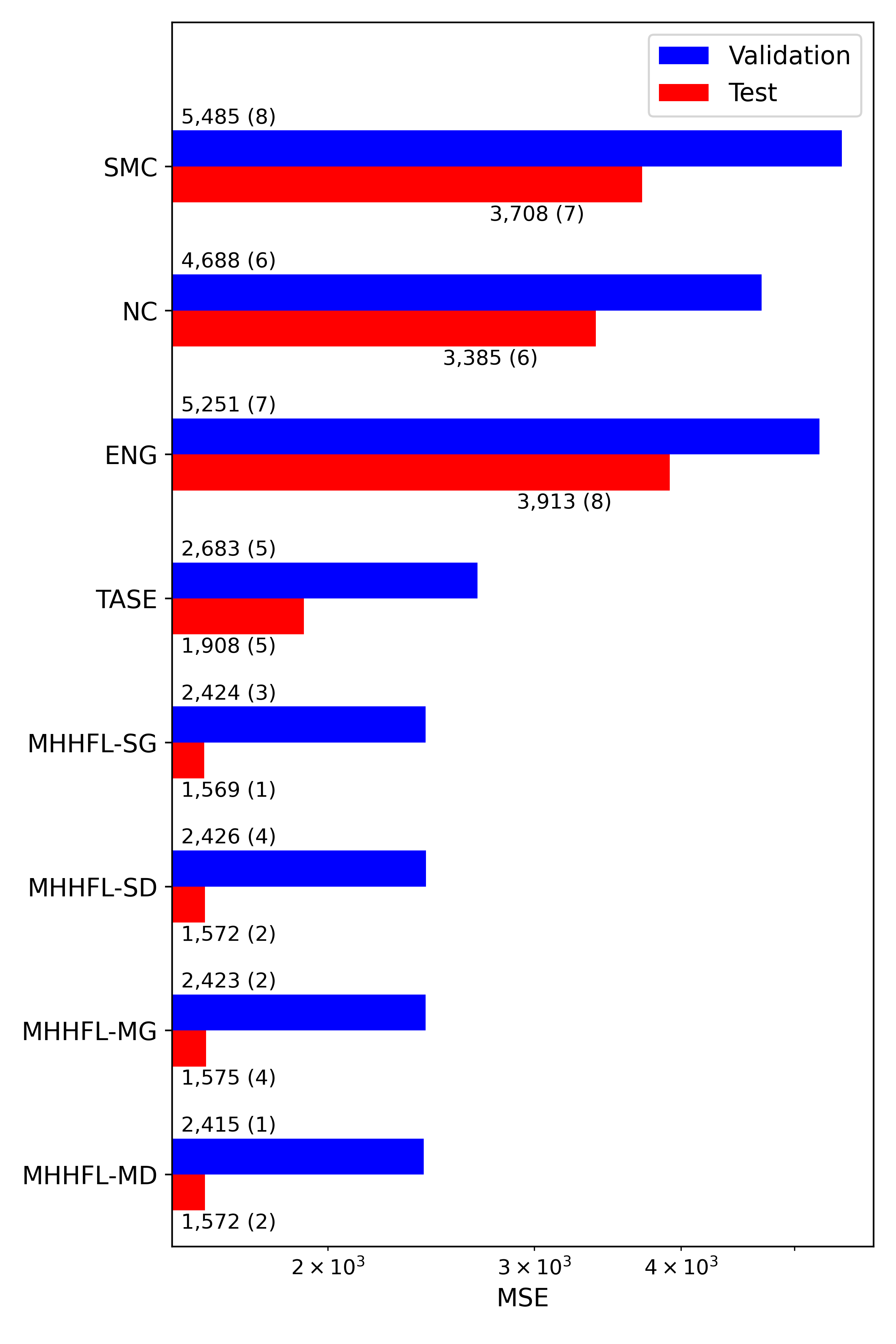}} 
    \subfigure[HuskyC]{\includegraphics[width=0.4\textwidth]{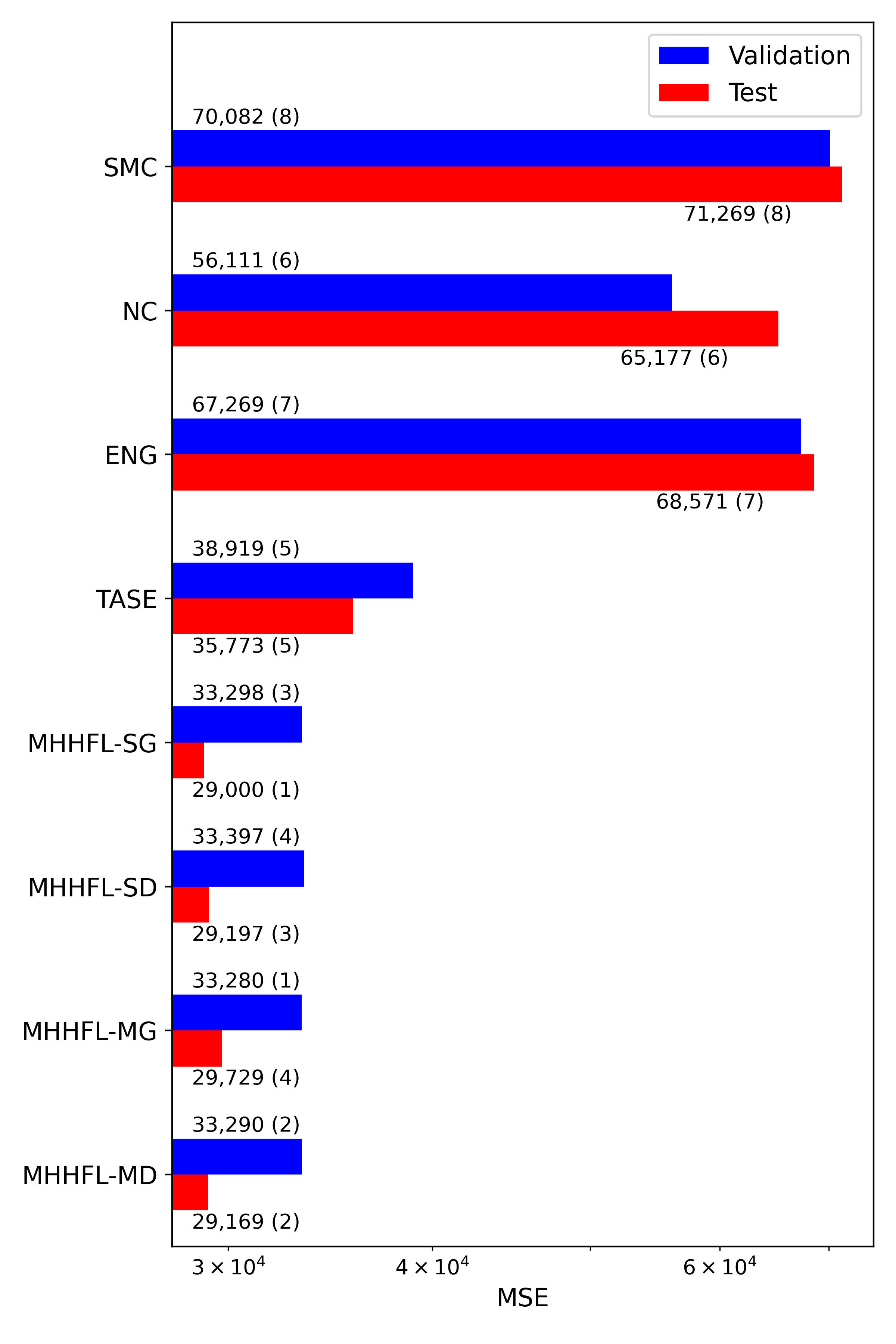}} 
    \caption{Robustness Evaluation}
    \label{fig:RE}
\end{figure}

\subsection{Robustness Evaluation}     \label{sec:Exp-RO}
Due to the uncertain environment, data collection might be delayed by various factors. Therefore, we evaluate robustness by predicting power consumption after 5 time steps (obtaining a buffer time of 0.5 seconds).
In other words, the systems use $X_{t}$ to predict $y_{t}$ in Section~\ref{sec:Exp-PE}, but use $X_{t}$ to predict $y_{t+4}$ in this section.
The experimental results of robustness evaluation are presented in Figure~\ref{fig:RE}, which obtain similar results in Figure~\ref{fig:RE}.
The blue and red bars represent the validation and testing MSE, respectively, and the values in parentheses represent the rankings among the systems.

The proposed MHHFL systems (MHHFL-SG, -SD, -MG, and -MD) always achieve the lowest validation error on the three Husky datasets, but are inferior to ENG~\cite{ENG} on the AIUT dataset.
However, all benchmark systems suffer from severe overfitting on the AIUT dataset, especially on ENG~\cite{ENG}, which has a low validation error of 1,296 but an extremely high testing error of 446,395.
The phenomenon exhibits weaknesses of benchmark systems (existing methods), and overfitting may occur due to using only a single data source.
As for the proposed MHHFL systems, they acquire knowledge from various and heterogeneous domains, and have almost no overfitting phenomenon.

In the testing set, the four proposed MHHFL systems always achieve the lowest prediction error on all datasets.
Comparing the proposed MHHFL-MD with the SOTA system (TASE~\cite{TASE}), MHHFL-MD reduces the MSE of the testing by 94.5\% (1 - 28,534 / 514,264) for AIUT, 4.9\% (1 - 3,101 / 3,261) for HuskyA, 17.6\% (1 - 1,572 / 1,908) for HuskyB, and 18.5\% (1 - 29,169 / 35,773) for HuskyC.
In summary, all the MHHFL systems significantly outperform the benchmark systems and reduce the prediction error of SOTA by 4.9\% to 94.5\%, which shows the effectiveness and robustness of the proposed MHHFL systems in the delayed scenario.

\begin{figure}[th]
    \centering
    \subfigure[AIUT]{\includegraphics[width=0.475\textwidth]{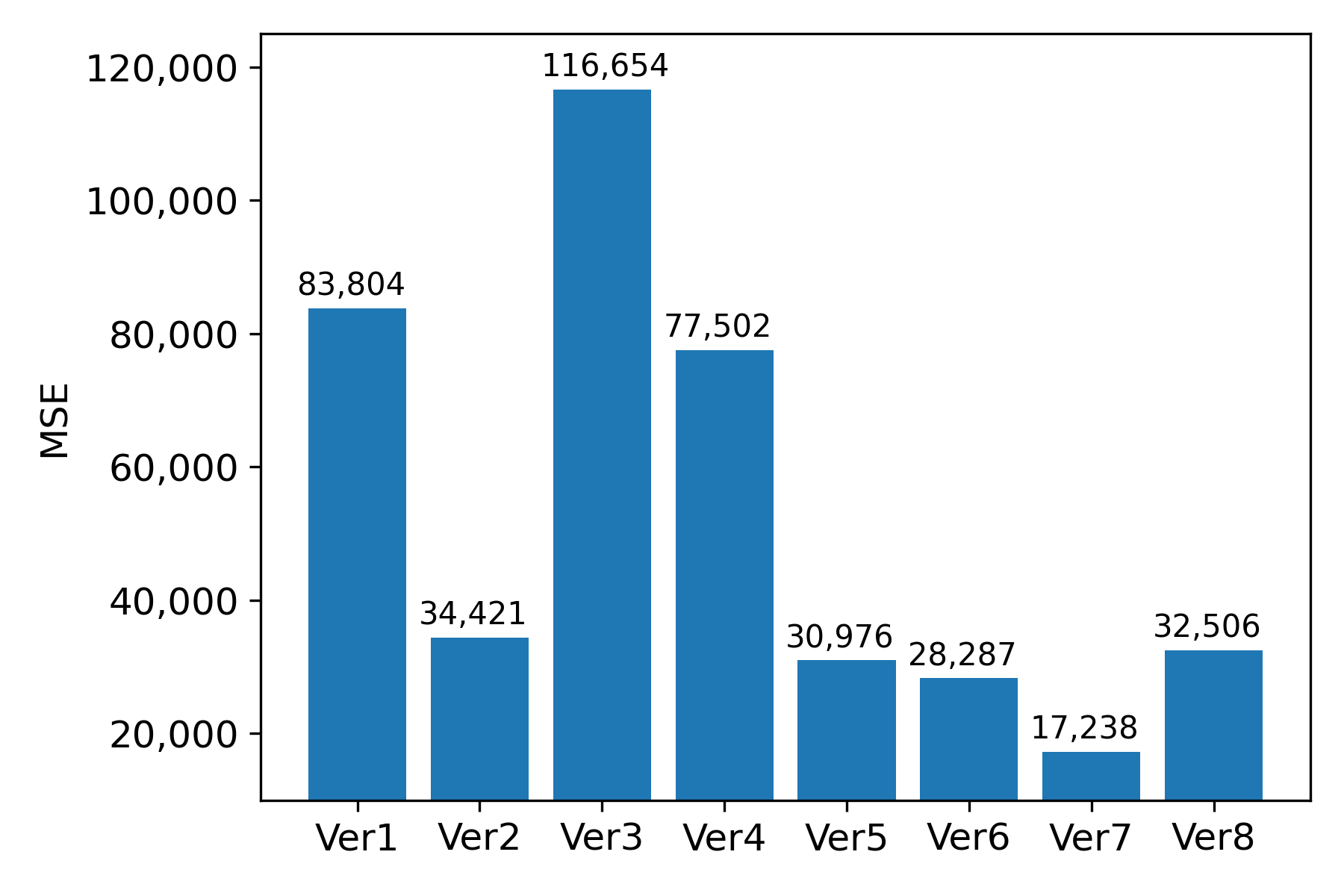}} 
    \subfigure[HuskyA]{\includegraphics[width=0.475\textwidth]{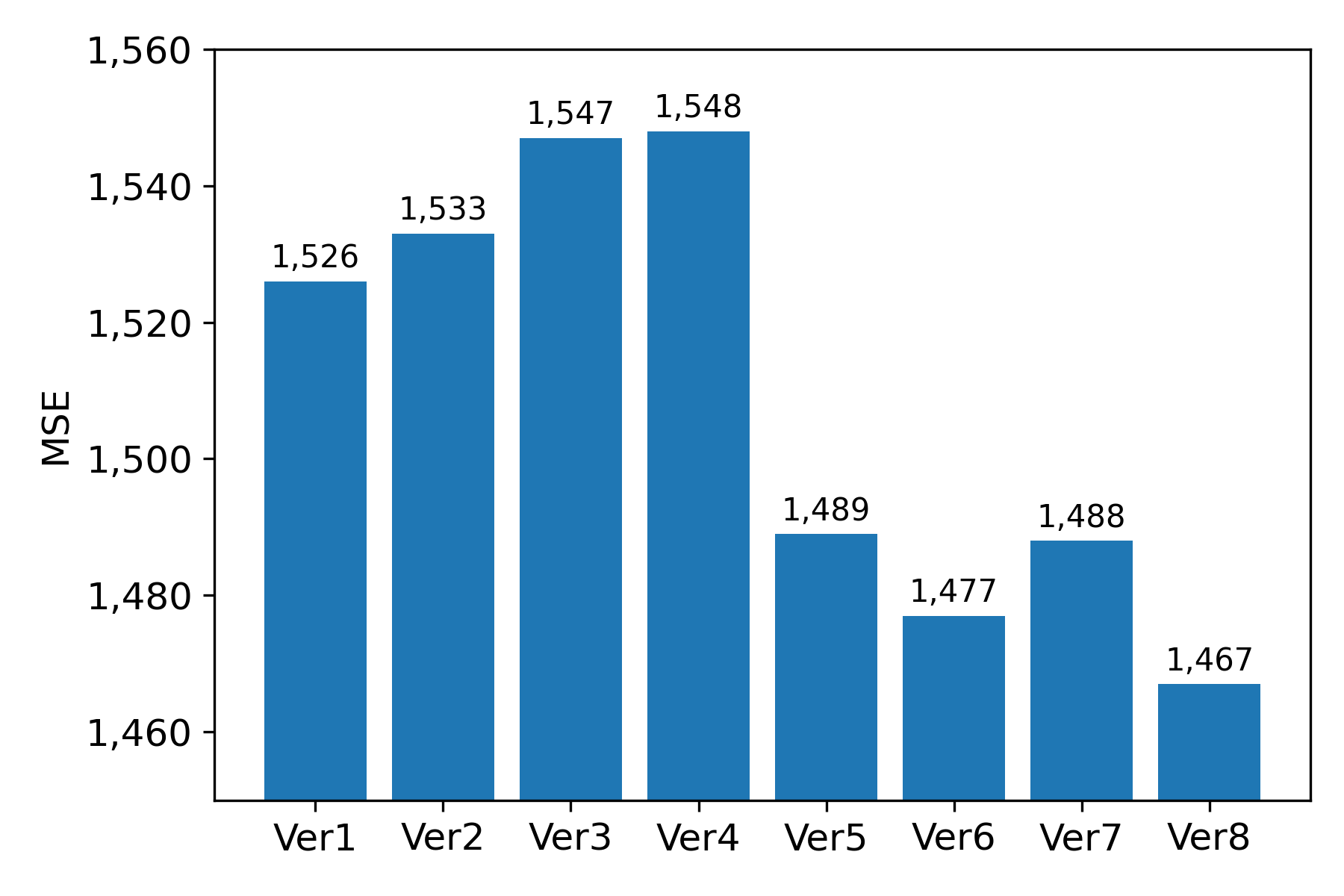}} 
    \centering
    \subfigure[HuskyB]{\includegraphics[width=0.475\textwidth]{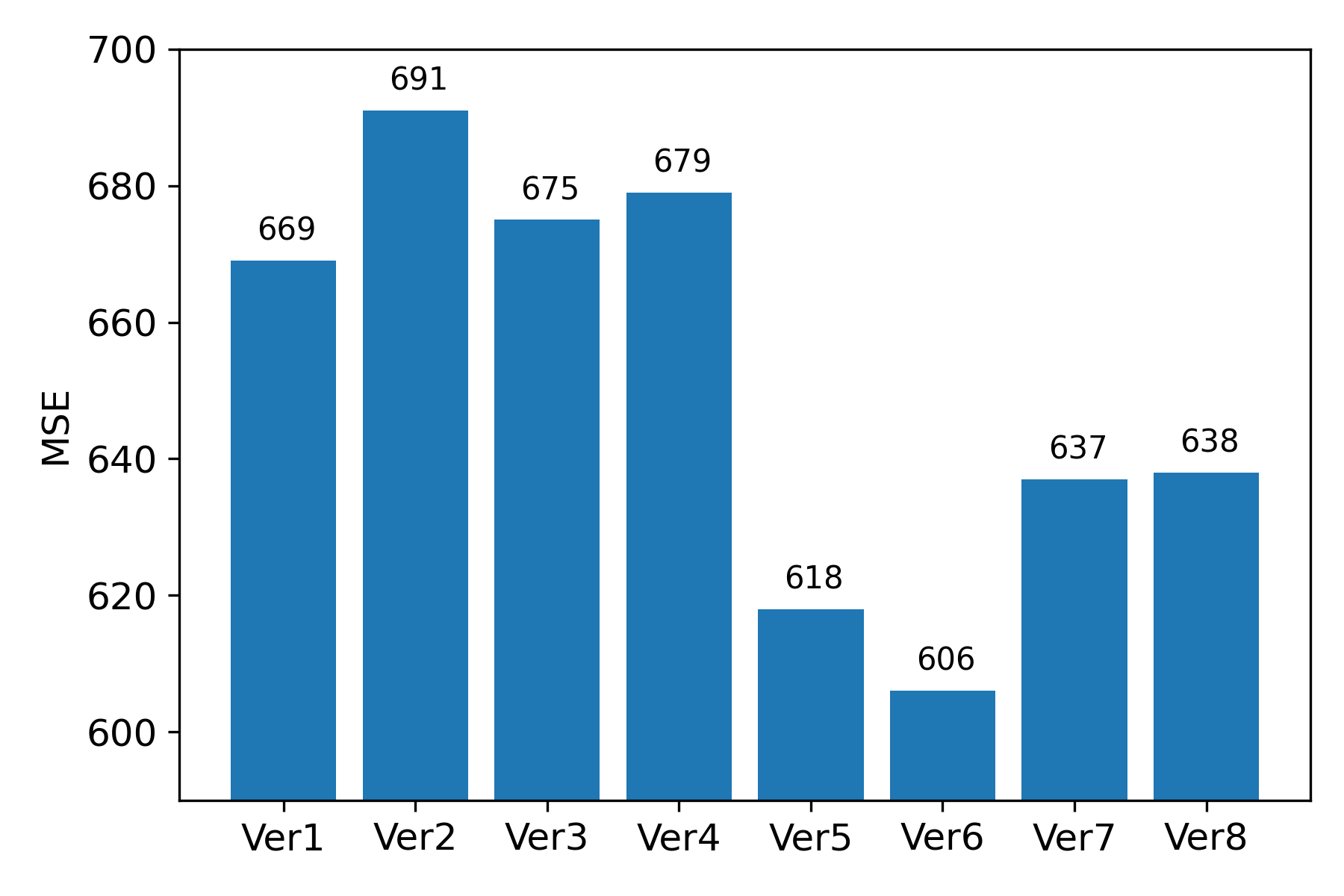}} 
    \subfigure[HuskyC]{\includegraphics[width=0.475\textwidth]{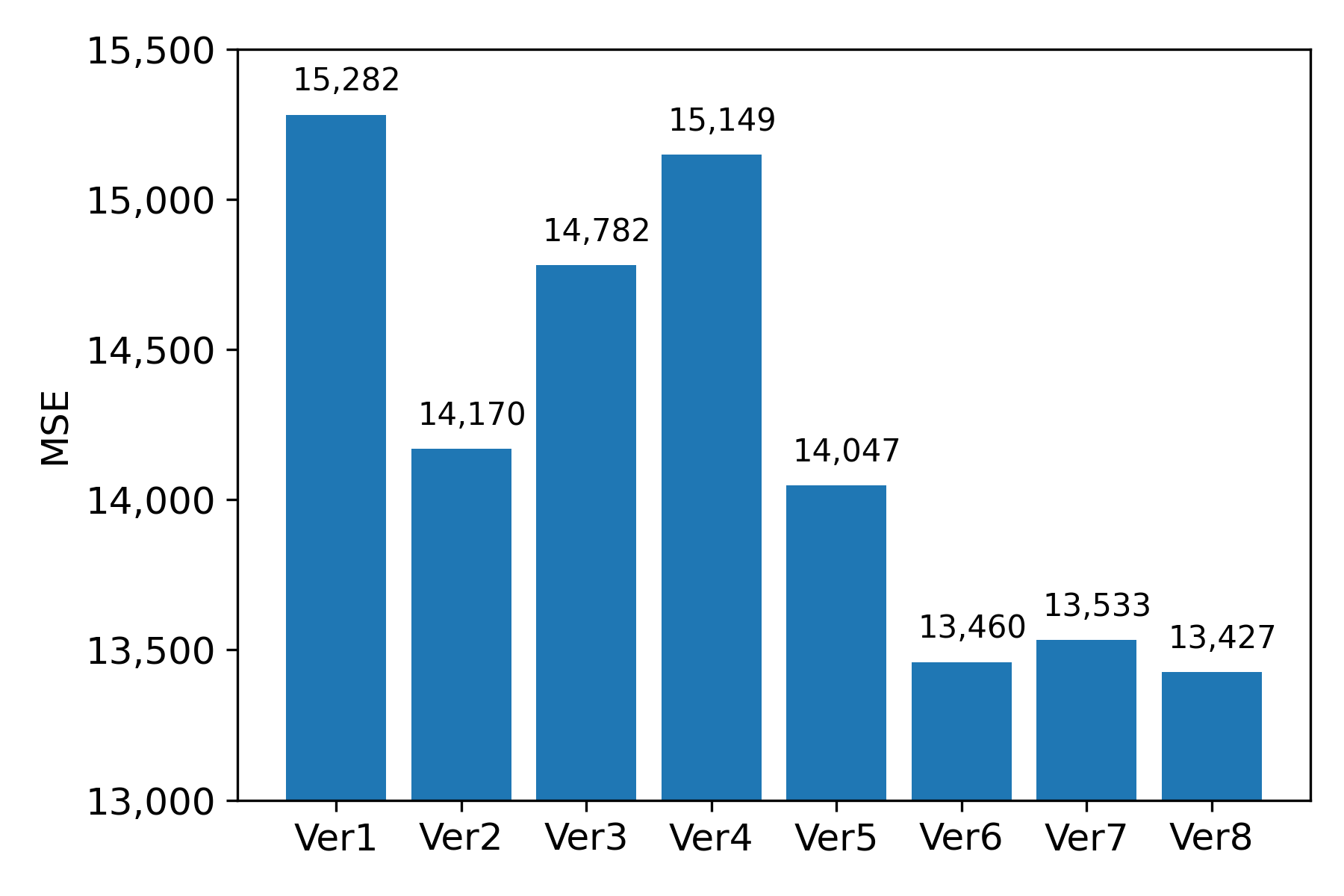}} 
    \caption{Ablation studies on four datasets}
    \label{fig:AS}
\end{figure}

\subsection{Ablation Studies}     \label{sec:Exp-AS}
In this section, we perform ablation studies to evaluate the effectiveness of each proposed mechanism, and we compare the performance of eight versions of MHHFL (Ver1 to Ver8).
Ver5, Ver6, Ver7, and Ver8 refer to MHHFL-SG, MHHFL-SD, MHHFL-MG, and MHHFL-MD, respectively.
Ver4 refers to MHHFL-S without head network selection, but with random selection.
Ver3 refers to MHHFL without head network selection but with traditional FedAvg~\cite{FedAvg} (average of weights of all head networks).
Ver2 denotes MHHFL without federated learning, and Ver1 denotes Ver2 without the pre-classification mechanism (defined in Seciton~\ref{sec:M-DE}).
The testing MSEs of eight versions of MHHFL on four datasets are shown in Figure~\ref{fig:AS}.

Compared with Ver1 and Ver2, the pre-classification mechanism (Ver2) slightly increases the MSE for HuskyA and HuskyB datasets (by 0.5\% and 3.3\%, respectively), but significantly reduces the MSE for AIUT and HuskyC datasets (by 58.9\% and 7.3\%, respectively), showing the partial effectiveness of the proposed pre-classification mechanism.
Compared with Ver2 and Ver3, the traditional FedAvg (Ver3) only slightly reduces the MSE at HuskyB by 2.3\%, but dramatically increases the MSE at the other datasets (238.9\%, 0.9\%, and 4.3\%, respectively), showing that the traditional FedAvg is not effective in the given scenario.
Compared to Ver2 and Ver4, heterogeneous federated learning (Ver4) with random selection slightly reduces the MSE on HuskyB by 1.7\%, but dramatically increases the MSE on the other datasets (125.2\%, 1.0\%, and 6.9\%).
Random selection (Ver4) actually performs worse than FedAvg (Ver3) in most cases, demonstrating the barrier of heterogeneity to knowledge transfer.

Compared to Ver4 and Ver5, the proposed heterogeneous federated learning (Ver5) significantly reduces the MSE of the tests by 60.0\%, 3.8\%, 9.0\%, and 7.3\%, respectively.
This phenomenon proves the effectiveness of the proposed heterogeneous federated learning, especially the head network embedding and selection mechanisms (compared to the random selection mechanism, Ver4).
As for the comparison between Ver5 and Ver8, there is no obvious and consistent advantage for single selection (Ver5 and Ver6), multiple selection (Ver7 and Ver8), gradient-based embedding (Ver5 and Ver7), and data-based embedding (Ver6 and Ver8).
The only finding is that the gradient-based embedding can always reduce prediction errors in MHHFL with the single selection mechanism.

In summary, the ablation studies demonstrate the effectiveness of the proposed pre-classification, head network embedding, and selection mechanisms in providing heterogeneous federated learning and significantly reducing the prediction error in most datasets.

\begin{figure}[th]
    \centering
    \subfigure[AIUT]{\includegraphics[width=0.475\textwidth]{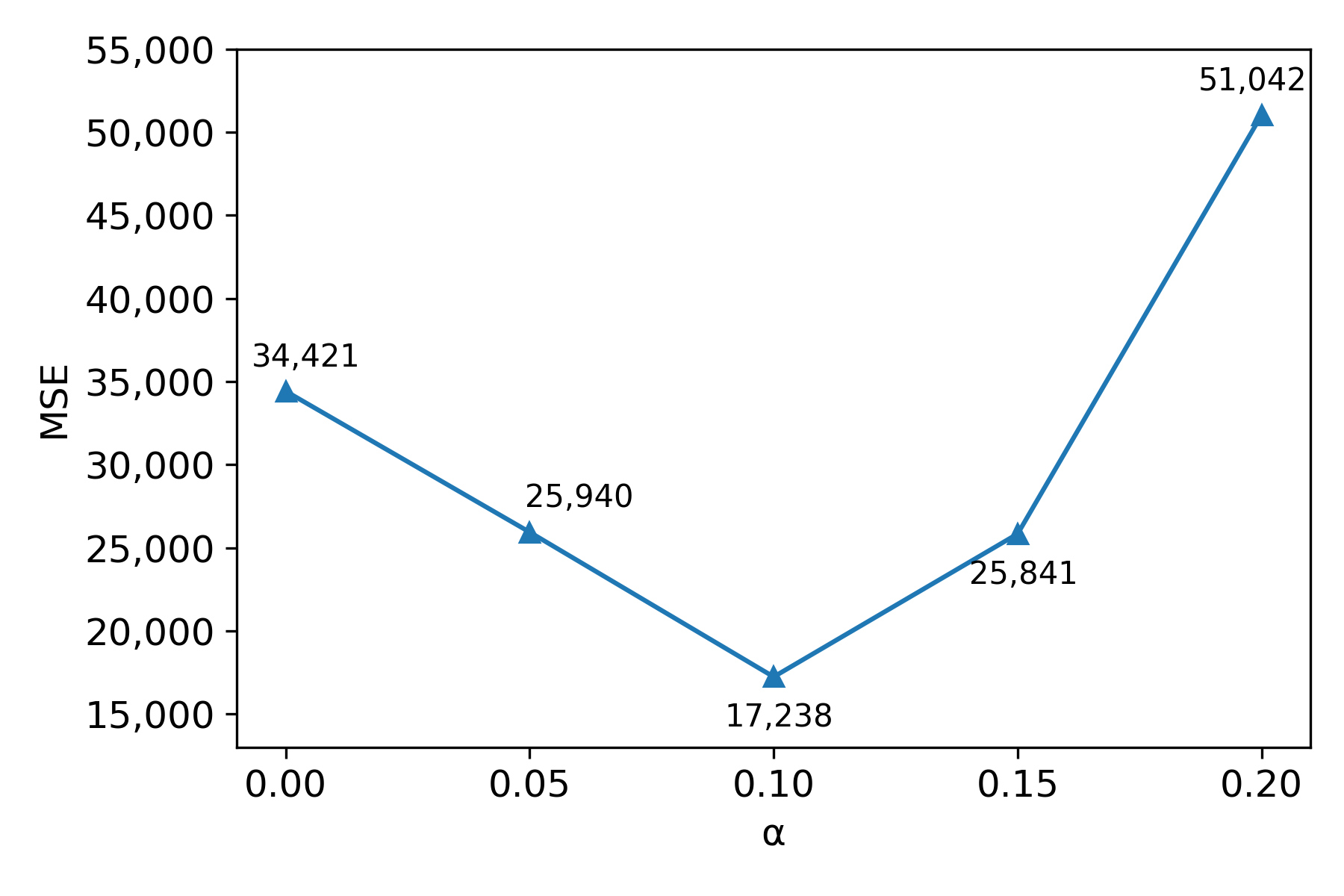}} 
    \subfigure[HuskyA]{\includegraphics[width=0.475\textwidth]{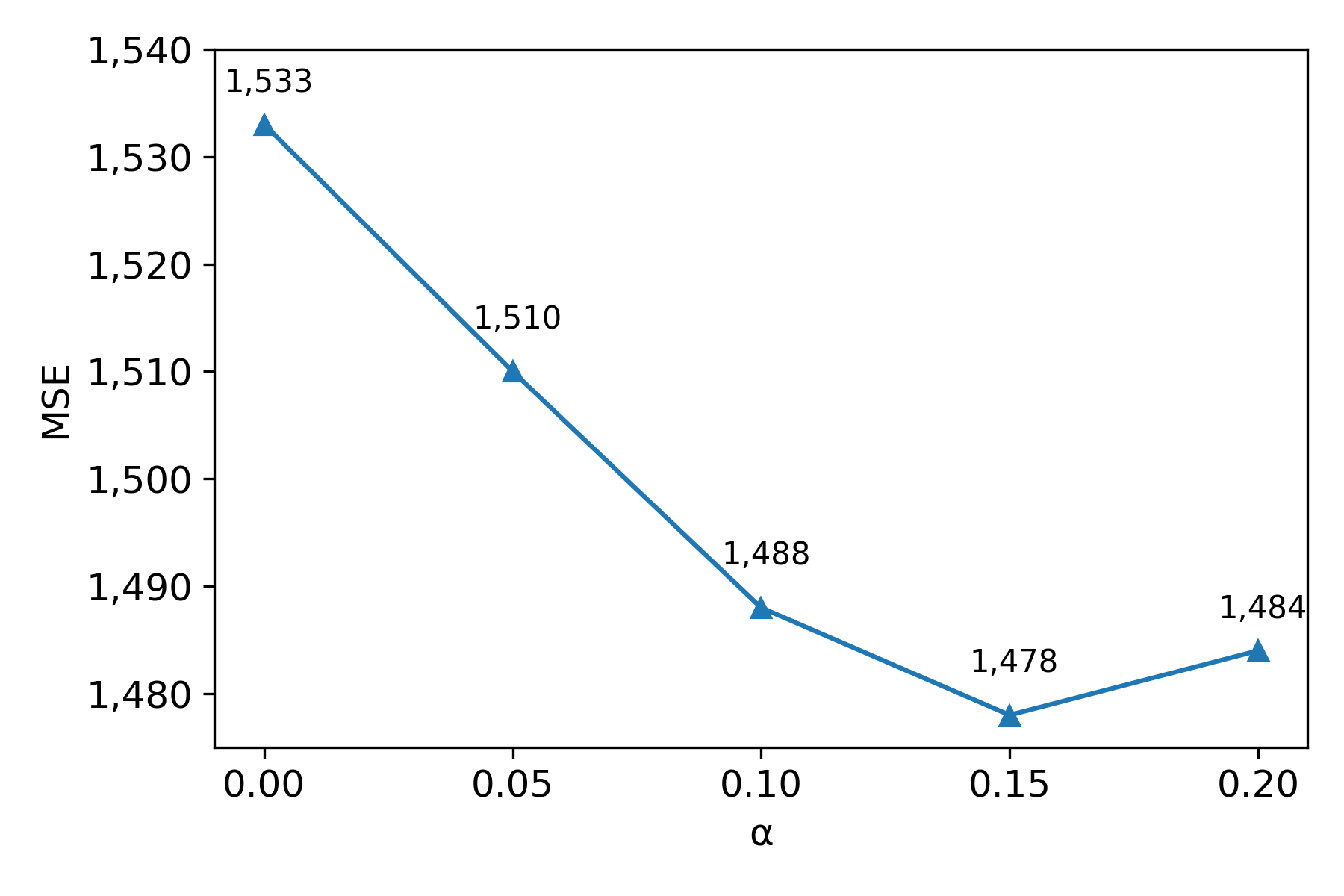}} 
    \centering
    \subfigure[HuskyB]{\includegraphics[width=0.475\textwidth]{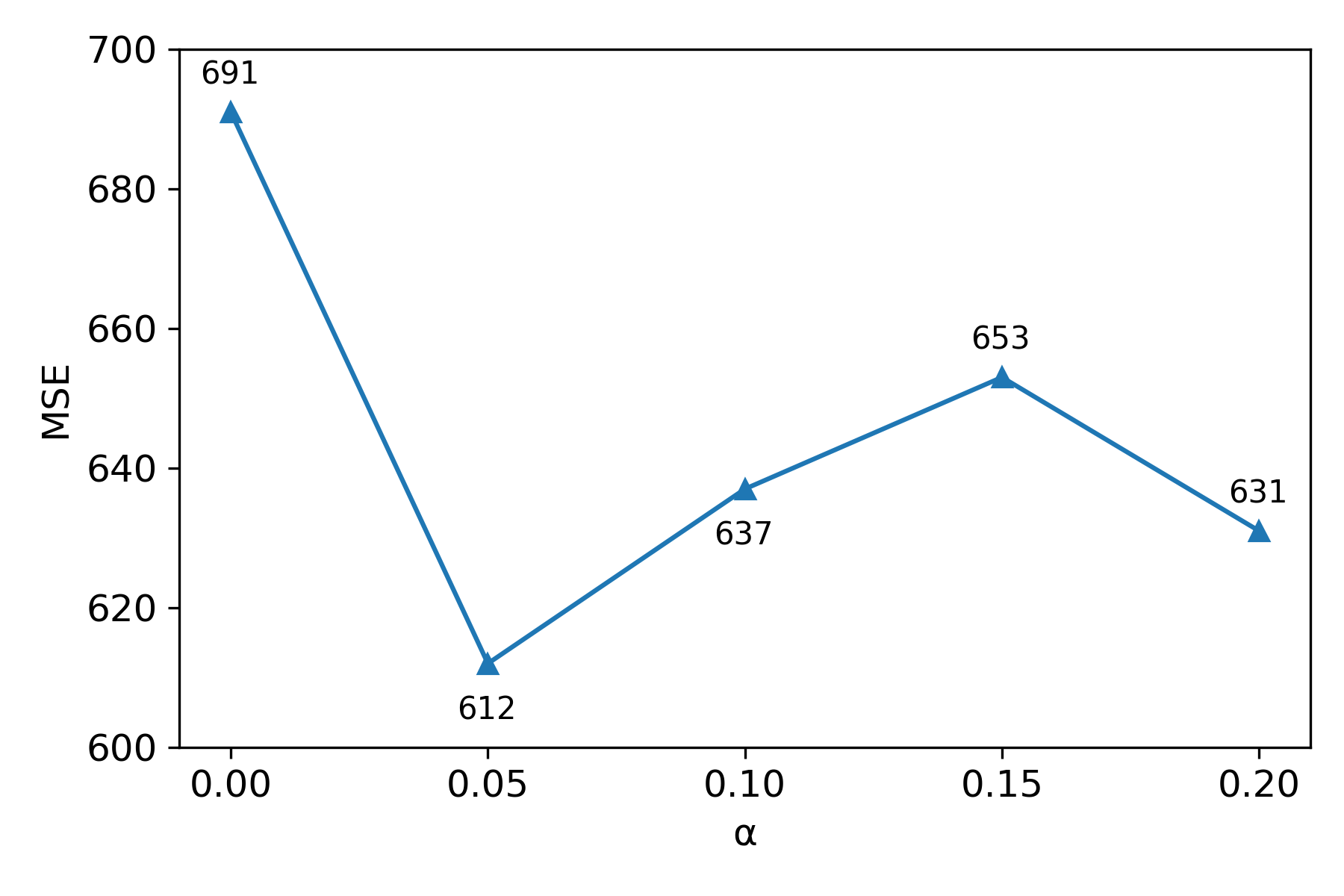}} 
    \subfigure[HuskyC]{\includegraphics[width=0.475\textwidth]{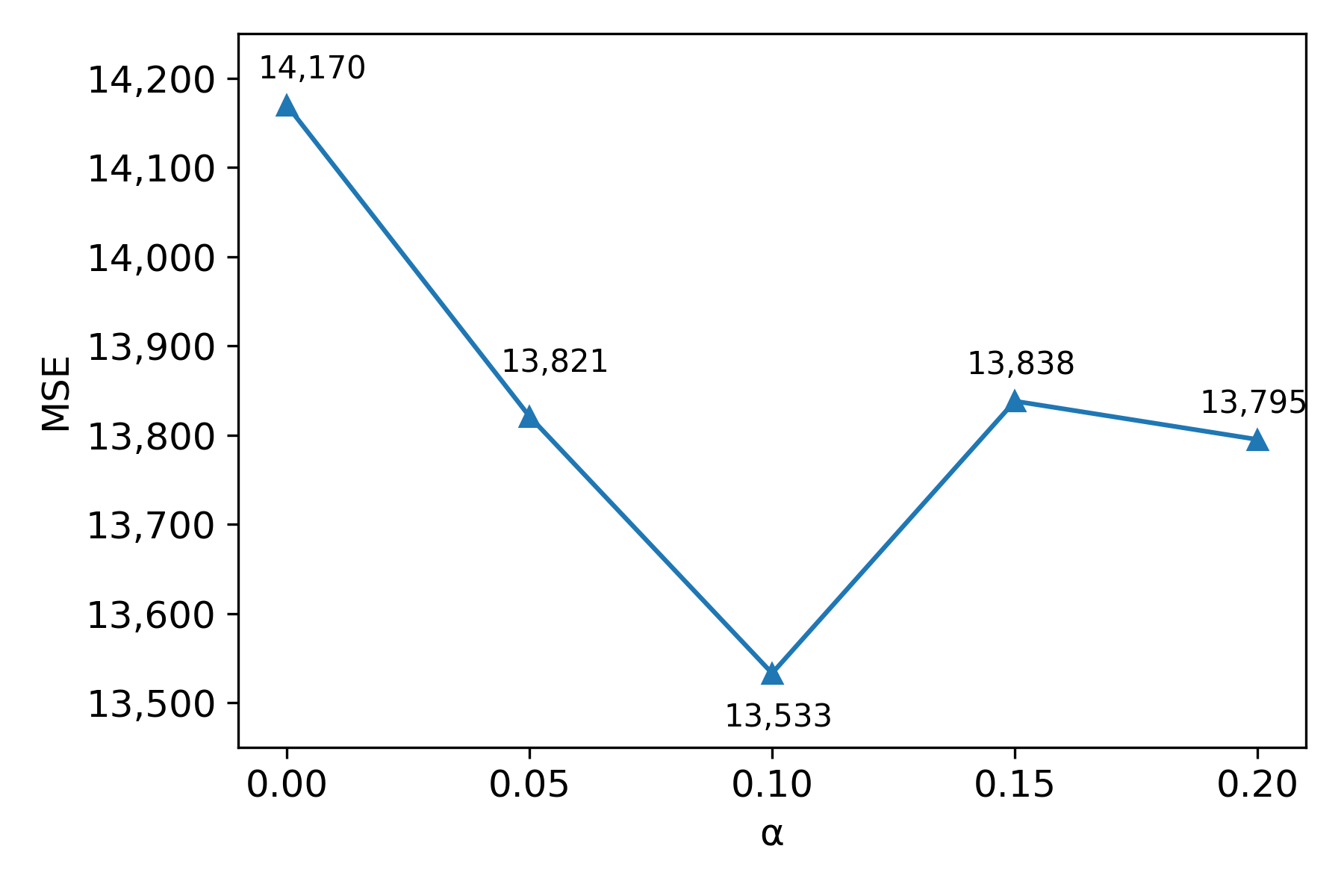}} 
    \caption{Sensitivity analysis on blending scale $\alpha$}
    \label{fig:SA}
\end{figure}

\subsection{Sensitivity Analysis}     \label{sec:Exp-SA}
The above experiments exhibit the improvement of heterogeneous federated learning, which blends the weights of the head networks to facilitate knowledge sharing.
To investigate the influence of blending scale $\alpha$ on the MHHFL, we perform the sensitivity analysis on the proposed MHHFL-MG, as shown in Figure~\ref{fig:SA}.
The x-axis of Figure~\ref{fig:SA} represents the blending scale $\alpha$ from 0.00 (neither blending nor federated learning) to 0.20, and the y-axis shows the testing MSE.

Most MSEs for $\alpha > 0$ are smaller than those for $\alpha = 0$ (the only exception is $\alpha = 0.2$ on the AIUT dataset).
The phenomenon exhibits the effectiveness of the proposed heterogeneous federated learning, which is better than no federated learning.
Moreover, it is noticeable that there are `V' shapes in all subfigures in Figure~\ref{fig:SA}, and the lowest points are around $\alpha$ from 0.05 to 0.15.
On the one hand, insufficient $\alpha$ provides inadequate (scale) knowledge transfer from federated learning, reduces effectiveness, and increases the MSE of the testing.
On the other hand, too high $\alpha$ provides over-much knowledge transfer from federated learning, reduces the fit to the local data, resulting in a general but not specific network for prediction, and also increases the MSE of the testing.
Therefore, a suitable blending scale $\alpha$ can further increase the prediction performance, and we recommend an $\alpha$ of 0.10.

\section{Conclusion}     \label{sec:Con}
Electric vehicles are becoming increasingly popular, leading to a growing need for power consumption prediction to optimize battery power and minimize energy waste.
However, vehicle power consumption prediction faces several challenges, such as complex driving conditions and limited energy storage.
To overcome these challenges, researchers have used various techniques, but existing models lack discussion of collaborative learning and privacy issues among multiple clients.

To address these issues, we propose a Multi-Head Heterogeneous Federated Learning (MHHFL) system that uses a data preprocessing module for pre-classifying features to facilitate the heterogeneous transfer.
The MHHFL system consists of multiple head networks and a prediction network, where each head network independently classifies a feature set and serves as a carrier for federated learning.
In the federated phase, the head network embedding mechanism embeds each head network into 2D vectors, and clients share the information (weights and embedded vectors) of the head networks with a central source pool.
The proposed selection mechanism then selects appropriate source networks based on the embedded vectors and mixes the target and source networks as knowledge transfer in federated learning.
Compared with traditional federated learning algorithms, the proposed MHHFL is characterized by its heterogeneity, asynchrony, efficiency, privacy, and security.

The experimental results show that the proposed MHHFL systems significantly outperform the benchmark systems and reduce the prediction error by 24.9\% to 94.1\%.
Robustness evaluation is performed to simulate the communication latency, and the MHHFL systems maintain their excellent performance and suppress the SOTA by 4.9\% to 94.5\%.
The ablation studies also demonstrate the effectiveness of the proposed mechanisms, especially heterogeneous federated learning (head network embedding and selection mechanisms), which significantly outperforms traditional FedAvg and random transfer.
In the future, we will develop novel pre-classification mechanisms that simultaneously enable anomaly detection and efficiently embed the networks.
In future work, we will evaluate the effectiveness of the proposed systems across various domains.

\section*{Acknowledgment}
This work is partially supported by the National Centre for Research and Development under the project Automated Guided Vehicles integrated with Collaborative Robots for Smart Industry Perspective, and the Project Contract no. is: NOR/POLNOR/CoBotAGV/0027/2019-00.

\bibliographystyle{elsarticle-num}
\bibliography{refs}
\end{document}